\documentclass[10pt,twocolumn,letterpaper]{article}

\usepackage{wacv}
\usepackage{times}
\usepackage{epsfig}
\usepackage{graphicx}
\usepackage{amsmath}
\usepackage{amssymb}

\usepackage{ntheorem}
\usepackage{url}
\usepackage{float}
\usepackage[misc]{ifsym}
\usepackage{comment}

\usepackage[pagebackref=true,breaklinks=true,letterpaper=true,colorlinks,bookmarks=false]{hyperref}

\wacvfinalcopy 

\def\wacvPaperID{244} 

\ifwacvfinal\fi
\begin{document}

\title{Learning a distance function with a Siamese network \\ to localize anomalies in videos}

\author{Bharathkumar Ramachandra\\
North Carolina State University\\
Raleigh, NC 27695\\
{\tt\small bramach2@ncsu.edu}
\and
Michael J. Jones\\
Mitsubishi Electric Research Labs (MERL)\\
201 Broadway, 8th floor; Cambridge, MA 02478\\
{\tt\small mjones@merl.com}
\and
Ranga Raju Vatsavai\\
North Carolina State University\\
Raleigh, NC 27695\\
{\tt\small rrvatsav@ncsu.edu}
}

\maketitle
\ifwacvfinal\thispagestyle{empty}\fi

\begin{abstract}
   This work introduces a new approach to localize anomalies in surveillance video.  The main novelty is the idea of using a Siamese convolutional neural network (CNN) to learn a distance function between a pair of video patches (spatio-temporal regions of video).  The learned distance function, which is not specific to the target video, is used to measure the distance between each video patch in the testing video and the video patches found in normal training video. If a testing video patch is not similar to any normal video patch then it must be anomalous.  We compare our approach to previously published algorithms using 4 evaluation measures and 3 challenging target benchmark datasets. Experiments show that our approach either surpasses or performs comparably to current state-of-the-art methods.
\end{abstract}

\section{Introduction}
\label{sect:intro}
Video anomaly detection is the task of localizing (spatially and
temporally) anomalies in videos, where anomalies refer simply to
unusual activity.  Unusual activity is scene dependent;  what is
unusual in one scene may be normal in another.  In order to define
what is normal, video of normal activity from the scene is provided.
In the formulation of video anomaly detection that we focus on in this
paper, we assume both the normal training video as well as the testing
video come from the same single fixed camera, the most common surveillance setting.
In this application, normal video (i.e. not containing any anomalies)
is simple to gather while anomalous video is not.  This is why it makes
sense to provide normal video (and only normal video) for training. 
Given this formulation, the problem becomes one of building a model of
normal activity from the normal training video and then detecting large
deviations from the model in testing video of the same scene as
anomalous.


Most previous methods have limitations that can be attributed to one
or more of the following, which serve as the motivation for our
approach: (1) The features used in many methods are hand-crafted.
Examples include spatio-temporal gradients \cite{lu_abnormal_2013},
dynamic textures \cite{mahadevan_anomaly_2010,weixin_li_anomaly_2014},
histogram of gradients \cite{hasan_learning_2016}, histogram of flows
\cite{hasan_learning_2016,saligrama_video_2012,cong_abnormal_2013},
flow fields \cite{adam_robust_2008,wu_chaotic_2010,mehran_abnormal_2009,antic_video_2011,antic_spatio-temporal_2015}
and foreground masks \cite{ramachandra_street_2019}.  (2) Almost
every method requires a computationally expensive model building phase
requiring expert knowledge which may not be practical for real
applications. (3) Many previous works focus on detecting only specific
deviations from normality as anomalous.

\begin{figure*}[ht]
\centering
\includegraphics[width=0.94\linewidth]{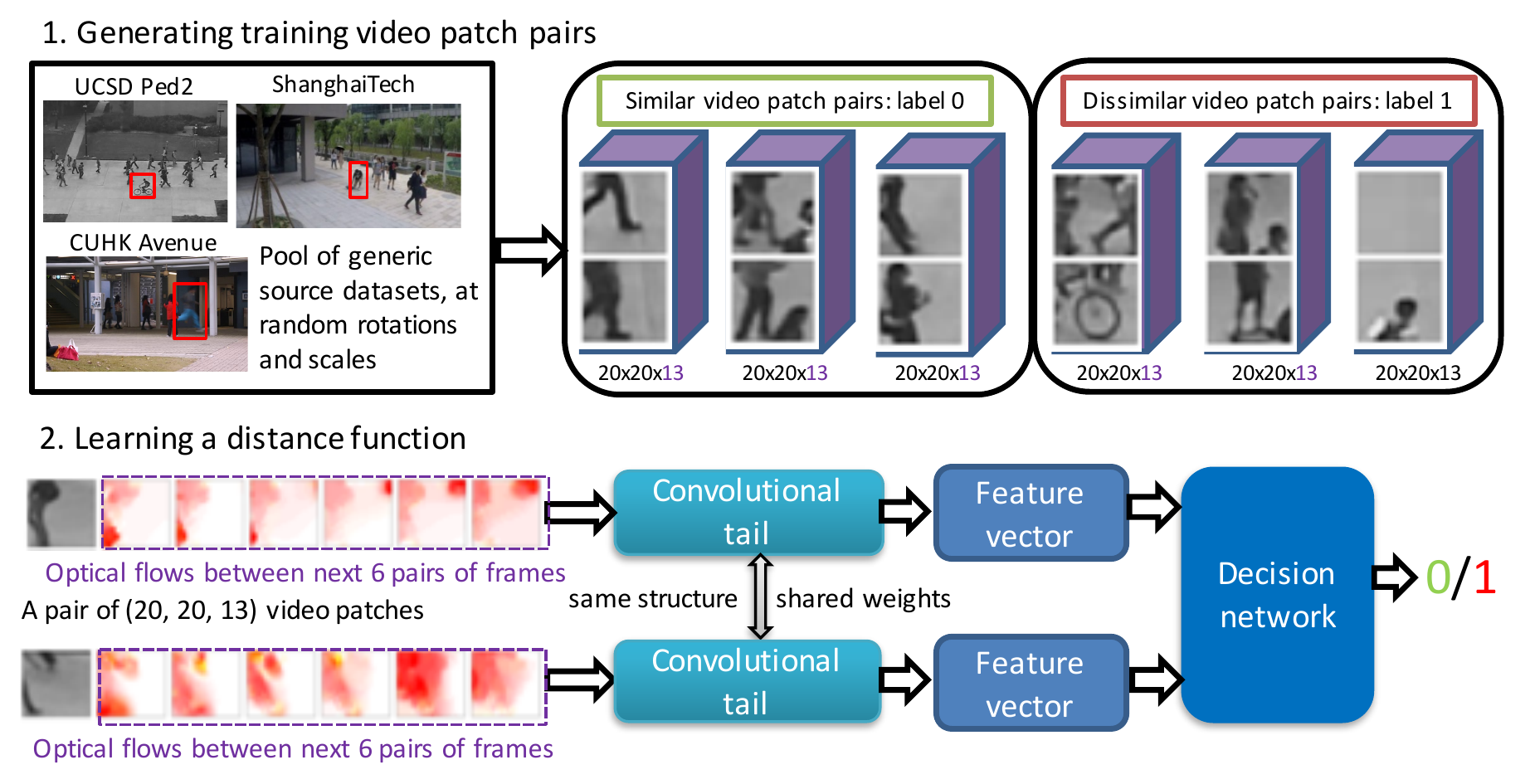}
\caption{An illustration of the scenario where UCSD Ped2, ShanghaiTech and CUHK Avenue are used as source datasets to learn a distance function from. Best viewed in color.}
\vspace{-10pt}
\label{fig:overview_1}
\end{figure*}

To overcome these limitations, we propose an exemplar-based nearest
neighbor approach to video anomaly detection that uses a distance
function learned by a Siamese CNN to measure how similar activity in
testing video is to normal activity.  Our approach builds on the work
of \cite{ramachandra_street_2019}, in which normal video is used to
create a model of normal activity consisting of a set of exemplars for
each spatial region of the video.  An exemplar is a feature vector
representing a video patch, i.e., a spatio-temporal block of video of
fixed size $H \times W \times T$ where $H$, $W$ and $T$ are the height,
width and temporal depth of a video patch \cite{dollar2005behavior}. 
The exemplars for a spatial region of video
represent all of the {\em unique} video patches that occur in the
normal video in that region. Exemplars are region-specific because
of the simple fact that anomalies are region-specific. To detect anomalies,
video patches from a particular spatial region in testing video are
compared against the exemplars for that region, and the anomaly score
is the distance to the nearest exemplar. If a testing video patch is
dissimilar to every exemplar video patch, then it is anomalous.  In
\cite{ramachandra_street_2019}, hand-crafted features (either
foreground masks or flow fields) were used to represent video patches
and a pre-defined distance function (either $L_2$ or normalized $L_1$)
was used to compute distances between feature vectors. 
We propose \textit{learning} a better feature vector and distance function by
training a Siamese CNN to measure the distance between pairs of video
patches. Our CNN is not specific to a particular scene, but is
trained from video patches from several different source video anomaly
detection datasets. This idea is similar in spirit to the work on
learning a CNN for matching patches
\cite{han_matchnet_2015,zagoruyko_learning_2015}, except extended to
video. Experiments show that our method either surpasses or performs comparably to the 
current state of the art on the UCSD Ped1, Ped2 \cite{mahadevan_anomaly_2010}
and CUHK Avenue \cite{lu_abnormal_2013} test sets.

In summary, our major contributions are:

1. Our approach transforms the problem of training a CNN to classify video patches as normal or anomalous (which cannot be done since we have no anomalous training examples) to the problem of training a CNN that computes the distance between two video patches (a problem for which we can generate plenty of examples). We use the \textit{same parameters for training the CNN from source datasets regardless of the target dataset}.

2. This approach allows task-specific feature learning, allows for efficient exemplar model building from normal video and detects a wide variety of deviations from normality as anomalous.

3. By shifting the complexity of the problem to the distance function learning task, the simple 1-NN distance-to-exemplar anomaly detection becomes highly interpretable. To the best of our knowledge, our paper is the first to take this approach to anomaly detection.

\begin{figure*}[ht]
\centering
\includegraphics[width=0.93\linewidth]{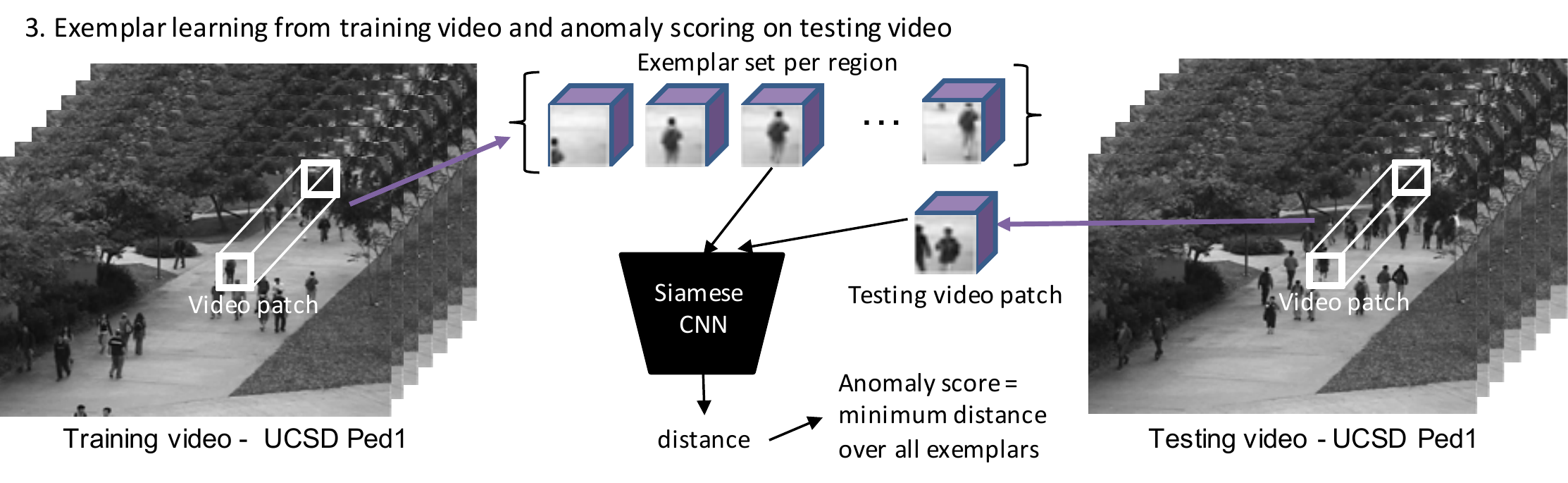}
\caption{An illustration of using the learned distance function to perform exemplar extraction and anomaly scoring on the target UCSD Ped1 dataset. Best viewed in color.}
\vspace{-10pt}
\label{fig:overview_2}
\end{figure*}

\section{Related Work}
\label{sect:rel-work}
Due to space constraints, we cannot do justice to the complete literature. We focus here on video anomaly detection methods that follow the formulation of the problem outlined previously. A number of methods such as \cite{ionescu_unmasking_2017,leibe_discriminative_2016,liu_classier_2018,sultani_real-world_2018} use other formulations of the video anomaly detection problem which we do not discuss here, although we organize this section similar to \cite{sultani_real-world_2018}. 

\subsection{Distance-based approaches}
\label{subsect:dist-based}
Distance-based approaches involve creating a model from a training partition and measuring deviations from this model to determine anomaly scores in the test partition.

The authors in \cite{saligrama_video_2012} use the insight that `optimal decision rules to determine local anomalies are local irrespective of normal behavior exhibiting statistical dependencies at the global scale' to collapse the large ambient data dimension. They propose local nearest neighbor based statistics to approximate these optimal decision rules to detect anomalies.

In \cite{xu_learning_2015}, stacked denoising auto-encoders are used
to learn both appearance and motion representations of video patches which are used with one-class SVMs to perform anomaly detection.

The authors in \cite{ravanbakhsh_plug-and-play_2018} derive an anomaly score map by consolidating the change in image features from a pre-trained CNN over the length of a video block.

\subsection{Probabilistic approaches}
\label{subsect:probabilistic}
Probabilistic approaches are similar to distance-based approaches, except that the model has a probabilistic interpretation, for example as a probabilistic graphical model or a high-dimensional probability distribution.

The authors in \cite{adam_robust_2008} use multiple fixed-location monitors to extract optical flow fields and compute the likelihood of an observation given the distribution stored in that monitor's buffer. 

In \cite{mahadevan_anomaly_2010}, the authors propose a representation comprising a mixture of dynamic textures (MDT), modeling a generative process for MDTs and discriminant saliency hypothesis test for anomaly detection. In \cite{weixin_li_anomaly_2014}, they build off the MDT representation to detect anomalies at multiple scales in a conditional random field framework.

Authors in \cite{antic_video_2011} contend that anomaly detection should try to ``explain away'' the normality in the test data using information learned from the training data. To this end, they use foreground object hypotheses and take a video parsing approach, treating those object hypotheses at test time which are necessary to explain the foreground but not explained by the exemplar training hypotheses are anomalous. In \cite{antic_spatio-temporal_2015}, they further build on this idea by extending the atomic unit of processing from an image patch to a video pipe.

\subsection{Reconstruction approaches}
\label{subsect:recon}
Reconstruction approaches aim to break down inputs into their common constituent pieces and put them back together to reconstruct the input, minimizing ``reconstruction error''.

\cite{hasan_learning_2016,chong_abnormal_2017,liu_future_2018,ravanbakhsh_abnormal_2017} are examples of methods that use this approach. In our experience, reconstruction based approaches seem to be naively biased against reconstructing faster motion, for the simple reason that absence of motion is much more common and easier to reconstruct.

A subset of reconstruction approaches, \textbf{sparse reconstruction approaches} have an additional constraint in that the reconstruction must be minimialistic, that is, using only a few essential features from a dictionary to perform the reconstruction. \cite{lu_abnormal_2013,luo_revisit_2017,cong_abnormal_2013} are examples of methods that use this approach.

Many of the methods mentioned above use deep networks.  All of the
previous papers that use deep networks for video anomaly detection
that we are aware of use them in one of two techniques: (1) either to
provide higher level features to represent video frames or (2) to
learn to reconstruct only normal video frames.  Much of the previous
work builds on the basic idea of using a CNN, either pre-trained on
image classification or other tasks
\cite{hinami_joint_2017,luo_revisit_2017,ravanbakhsh_plug-and-play_2018,smeureanu_deep_2017}
or trained on the training partitions of each video anomaly detection
dataset \cite{xu_learning_2015}, to provide a feature vector for
representing video frames.  The CNN feature maps provide higher level
features than raw pixels.  The other major theme of deep network
approaches is to learn an auto-encoder
\cite{hasan_learning_2016,chong_abnormal_2017} or generative
adversarial network \cite{ravanbakhsh_abnormal_2017,liu_future_2018}
to learn to reconstruct or predict only normal video frames.
Reconstruction error is then used as an anomaly score.  Our method
follows neither of these previous techniques and instead presents a
new way to take advantage of the power of deep networks for video
anomaly detection.  Namely, we use a CNN to learn a distance function
between pairs of video patches.  Thus, ours is a novel distance-based
approach.

\section{Method}
\label{sect:method}
By building on the exemplar-based nearest neighbor approach of
\cite{ramachandra_street_2019}, our main problem is to learn a
distance function for comparing video patches from testing video to
exemplar video patches that represent all of the unique video patches
found in the normal video.  To do this we use a Siamese network (see Figure \ref{fig:overview_1})
similar to the one first introduced by Bromley and LeCun
\cite{bromley_signature_1994}.  In essence, by making the anomaly
detection task itself a rather simple nearest neighbor distance
computation (see Figure \ref{fig:overview_2}), we seek to offload the
burden of modeling the complexity in this problem to the task of
learning a distance function. This learning problem can be done
offline and has a large amount of training data available from source
datasets. Ideally this can be done once and the resulting feature
representation and distance function used on a wide variety of 
different target datasets. 

In this section, we go into more detail in each of the steps shown in
Figures \ref{fig:overview_1} and \ref{fig:overview_2}, provide
justifications for our design decisions and setup some language
essential for the Experiments section.

\begin{figure*}[ht]
\centering
\includegraphics[width=\linewidth]{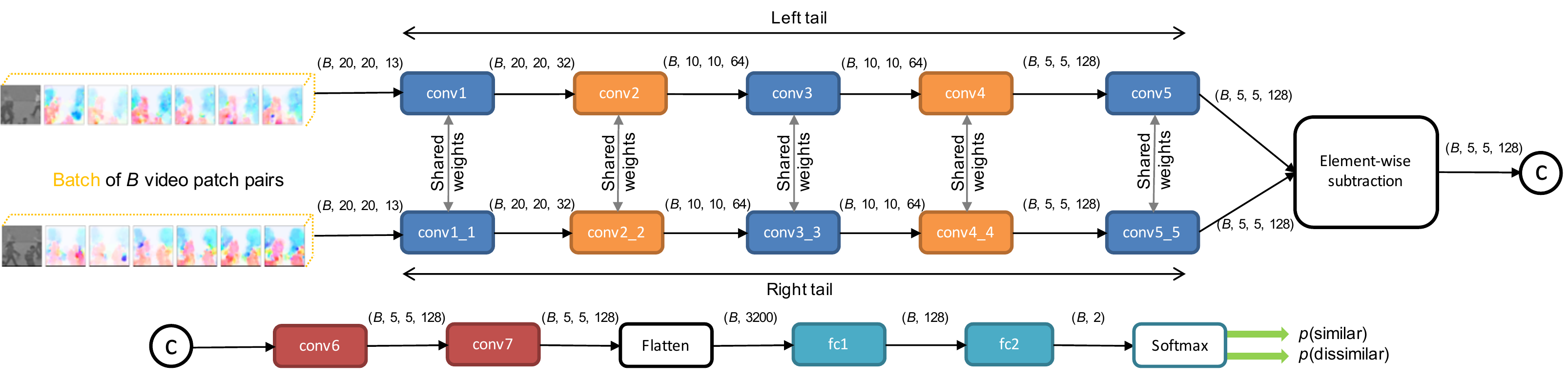}
\caption{Architecture of the Siamese neural network that learns a distance function between video patches. Best viewed in electronic form in color; color coding denotes unique structure.}
\label{fig:architecture}
\end{figure*}

\subsection{Generating training video patch pairs}
\label{subsect:dataset-curation} 
The main difficulty with training a Siamese network to estimate the
distance between a pair of video patches is determining how to
generate the training set of similar and dissimilar video patch pairs.
One training example consists of a pair of video patches plus a binary
label indicating whether the two video patches are similar or
dissimilar (see Figure \ref{fig:overview_1} part 1).  Video patch
pairs should be selected to correctly correspond to their ground truth
labels of ``similar'' or ``dissimilar''.  Pairs should also be picked
such that coverage of the possible domain of inputs to the CNN during
test time is high. This is to ensure that the CNN is not asked to
operate on out-of-domain inputs at test time.

How can we determine whether two video patches are similar or
dissimilar and how can we select a varied set of video patch pairs
that are relevant to video anomaly detection? An important insight is
that we can use existing video anomaly detection datasets to do this.
We use a source set of labeled video anomaly detection datasets to
generate similar and dissimilar video patch pairs.  The labeled
datasets used to generate training examples should of course be
disjoint from the target video anomaly detection dataset on which
testing will eventually be done.  The basic insight is as follows: for
each source dataset,

(1) A non-anomalous video patch from the test partition is similar to
at least one video patch from the same spatial region in the train
partition.  If it were not similar to any normal video patches it
would be anomalous.

(2) An anomalous video patch from the test partition is dissimilar to
all possible patches from the same spatial region in the train
partition.  Moreover, it is dissimilar to even the most similar video
patch.

The first rule generates a single pair for each normal video patch in
a test video, although since there are many normal video patches in
any test video, this rule can generate many similar pairs.  The second
rule generates many different dissimilar pairs for each anomalous
video patch in a test video.  The first rule requires a distance
function to find the most similar train video patch to a test video
patch.  It is also useful in the second rule to have a distance
function to know which dissimilar pairs are the most difficult
(i.e. similar) since these are the most useful for training. We use a
simple normalized L1 distance as our distance function along with the
representation of video patches described in Section
\ref{subsect:learning-dist}.

A reasonable concern about using a predefined distance function to
help select training examples is that the Siamese network might simply
learn this distance function.  This does not happen for a few reasons.
One is that the label for each example pair is not the L1 distance,
but rather a 0 or 1 indicating whether the pair is similar or
dissimilar, respectively.  Secondly, it is possible for the L1
distance between two similar pairs to be larger than the L1 distance
between two dissimilar pairs.

One important point to note is that normalized L1 distance is far from
ideal to measure distance between video patches. For example, this
distance does not take into account many variations in natural images
such as scale, illumination and pose of objects. Because these
variations mostly exist \textit{across different regions} in the
camera's field of view, we determine an adaptive threshold on
normalized L1 distance below which to perform these pairings. The
threshold for a region is determined by taking into account the above
rules in combination with inspecting the distribution of nearest
neighbor distances in a given region. Specifically, an adaptive
threshold for a given region in the camera frame is determined
simply as $\mu + \alpha * \sigma$ where $\mu$ is the mean of nearest
neighbor distances between testing video patches and training video
patches, $\sigma$ is the corresponding standard deviation and $\alpha$
is determined by identifying an elbow in the distribution of nearest
neighbor distances (we set it to 0.2 consistently in experiments). The adaptive threshold is common across the source
datasets but different for similar and dissimilar pairs. Notice that
dissimilar pairs that have large distances are more likely to be easy
to discriminate for the Siamese network; on the other hand, we require
some of these pairings despite this property to achieve high domain
converage. Thus, we include candidate pairs with
probability inversely proportional to the distance between them,
achieving high domain coverage, but also a sufficient number of examples
close to the decision boundary. We also include as similar pairs
random video patches paired with slightly augmented (random
translation and/or central scaling) versions of them. Our final video
patch pair dataset consists of an equal number of similar and
dissimilar pairs.

\subsection{Learning a distance function}
\label{subsect:learning-dist}
\textbf{Choice of representation:} At this point, it is important to choose how video patches are represented, such that the learned distance function will perform well in the anomaly detection task. Our choice of representation consists of a $H \times W \times C$ cuboid. In light of all anomalies being appearance or motion based, we adopt a multi-modal representation. In all our experiments that follow, the first channel is a grayscale image patch and the next 12 channels are image patches from absolute values of x and y directional gradients of dense optical flow fields (we use \cite{liu_beyond_nodate}) between the subsequent 6 pairs of image patches. This sets $C=13$ and we set $H=20$ and $W=20$ for all experiments. See Figure \ref{fig:overview_1} (part 2) for an illustration.

\textbf{Pre-processing:} Data augmentation of a random amount is performed on every video patch pair $x_1, x_2$ during training in order to improve the robustness of the learned distance function to these variations. The data augmentation involves randomly flipping left to right, centrally scaling in [0.7, 1] and brightness jittering of the first channel in [-0.2, 0.2] in a stochastic manner on both video patches in a pair. Pre-processing also involves linearly scaling intensity values of each video patch from [0, 255] to [-1, 1].

\textbf{Network architecture and training:} Figure \ref{fig:architecture} outlines our network architecture. Each video patch in a pair is first processed independently using conv-relu-batchnorm operations with $2 \times 2$ max-pooling after every other convolution in what we call convolutional twin ``tails''. Weight tying between the tails guarantees that two extremely similar video patches could not possibly have very different intermediate representations because each tail computes the same function. Finally, flattened feature vectors from the two twin tails (conv5, conv5\_5) are subtracted element-wise and processed consequently in a typical classification pipeline minimizing a cross-entropy loss. All convolutions use $3 \times 3$ filters with a stride of 1. We find that subtracting the feature maps at conv5 produces faster optimization when compared to concatenation. We think this is because element-wise subtraction induces a stronger structural prior on the network architecture. Let $B$ represent minibatch size, where $i$ indexes the minibatch and $\mathbf{y}(x_1^{(i)}, x_2^{(i)})$ be a length-$B$ vector which contains the labels for the mini-batch, where we assume $y(x_1^{(i)}, x_2^{(i)}) = 0$ whenever $x_1$ and $x_2$ are similar video patches and $y(x_1^{(i)}, x_2^{(i)}) = 1$ otherwise. The cross-entropy loss is of the form:

\begin{table*}[ht]
\centering
\resizebox{0.9\linewidth}{!}{%
\begin{tabular}{llllll}
\hline
\textbf{Method} & \textbf{\begin{tabular}[c]{@{}l@{}}UCSD Ped1\\ frame AUC/EER\end{tabular}} & \textbf{\begin{tabular}[c]{@{}l@{}}UCSD Ped1\\ pixel AUC*\end{tabular}} & \textbf{\begin{tabular}[c]{@{}l@{}}UCSD Ped2\\ frame AUC/EER\end{tabular}} & \textbf{\begin{tabular}[c]{@{}l@{}}UCSD Ped2\\ pixel AUC\end{tabular}} & \textbf{\begin{tabular}[c]{@{}l@{}}CUHK Avenue \\ frame AUC/EER\end{tabular}}     \\ \hline
Adam \cite{adam_robust_2008} & 65.0\%/38.0\% & 46.1\% & 63.0\%/42.0\% & 18.0\% & -/- \\ \hline
Social force \cite{mehran_abnormal_2009} & 67.5\%/31.0\% & 19.7\% & 63.0\%/42.0\% & 21.0\% & -/- \\ \hline
MPPCA \cite{mahadevan_anomaly_2010} & 59.0\%/40.0\% & 20.5\% & 77.0\%/30.0\%  & 14.0\% & -/- \\ \hline
Social force + MPPCA \cite{mahadevan_anomaly_2010} & 67.0\%/32.0\% & 21.3\% & 71.0\%/36.0\% & 21.0\% & -/- \\ \hline
MDT \cite{mahadevan_anomaly_2010} & 81.8\%/25.0\% & 44.1\% & 85.0\%/25.0\% & 44.0\% & -/- \\ \hline
AMDN \cite{xu_learning_2015} & 92.1\%/16.0\% & 67.2\% & 90.8\%/17.0\% & - & -/- \\ \hline
Video parsing \cite{antic_video_2011} & 91.0\%/18.0\% & 83.6\% & 92.0\%/14.0\% & 76.0\% & -/- \\ \hline
Local statistical aggregates \cite{saligrama_video_2012} & 92.7\%/16.0\% & - & -/- & - & -/- \\ \hline
Detection at 150 FPS \cite{lu_abnormal_2013} & 91.8\%/15.0\% & 63.8\% & -/- & - & -/- \\ \hline
Sparse reconstruction \cite{cong_abnormal_2013} & 86.0\%/19.0\% & 45.3\% & -/- & - & -/- \\ \hline
HMDT CRF \cite{weixin_li_anomaly_2014} & -/17.8\% & 82.7\% & -/18.5\% & - & -/- \\ \hline
ST video parsing \cite{antic_spatio-temporal_2015} & 93.9\%/12.9\% & \textbf{84.2\%} & 94.6\%/\textbf{10.6\%} & 81.1\% & -/- \\ \hline
Conv-AE \cite{hasan_learning_2016} & 81.0\%27.9\% & - & 90.0\%/21.7\% & - & 70.2\%/25.1\% \\ \hline
Deep event models \cite{feng_learning_2017} & 92.5\%/15.1\% & 69.9\% & -/- & - & -/- \\ \hline
Compact feature sets \cite{leyva_video_2017} & 82.0\%/21.1\% & 57.0\% & 84.0\%/19.2\% & - & -/- \\ \hline
Convex polytope ensembles \cite{turchini_convex_2017} & 78.2\%/24.0\% & 62.2\% & 80.7\%/19.0\% & - & -/- \\ \hline
Joint detection and recounting \cite{hinami_joint_2017} & -/- & - & 92.2\%/13.9\% & 89.1\% & -/- \\ \hline
Sparse coding revisit \cite{luo_revisit_2017} & -/- & - & 92.2\%/- & - & 81.7\%/- \\ \hline
GAN \cite{ravanbakhsh_abnormal_2017} & \textbf{97.4\%/8.0\%} & 70.3\% & 93.5\%/14.0\% & - & -/- \\ \hline
Future frame prediction \cite{liu_future_2018} & 83.1\%/- & - & 95.4\%/- & - & 85.1\%/- \\ \hline
Plug and play CNN \cite{ravanbakhsh_plug-and-play_2018} & 95.7\%/\textbf{8.0\%} & 64.5\% & 88.4\%/18.0\% & - & -/- \\ \hline
Narrowed normality clusters \cite{ionescu_detecting_2019} & -/- & - & -/- & - & 88.9\%/- \\ \hline
Object-centric auto-encoders \cite{ionescu_object-centric_2019} & -/- & - & \textbf{97.8\%}/- & - & \textbf{90.4}\%/- \\ \hline
NN on video patch FG masks \cite{ramachandra_street_2019} & 77.3\%/25.9\% & 69.3\% & 88.3\%/18.9\% & 83.9\% & 72.0\%/33.0\% \\ \hline 
\hline
\textbf{Ours} & 86.0\%/23.3\% &  80.4\% & 94.0\%/14.1\% & \textbf{93.0\%} & 87.2\%/\textbf{18.8\%}\\ \hline
\end{tabular}
}
\vspace{1pt}
\caption{Traditional frame-level and pixel-level evaluation criteria on the UCSD Ped1, UCSD Ped2 and CUHK Avenue benchmark datasets from related literature, ordered chronologically, complied from this same list. Our approach either surpasses or performs comparably on these evaluation criteria when compared to previous methods. *Some of the earlier works unfortunately use only a partially annotated subset available at the time to report performance.}
\vspace{-10pt}
\label{tab:auc-table}
\end{table*}

\begin{equation}
\begin{split}
\mathcal{L}(x_1^{(i)}, x_2^{(i)}) = - \gamma *  \mathbf{y}(x_1^{(i)}, x_2^{(i)}) \; \text{log} \; \mathbf{p}(x_1^{(i)}, x_2^{(i)}) \\ -  
(1 - \mathbf{y}(x_1^{(i)}, x_2^{(i)})) \; \text{log} \; (1 - \mathbf{p}(x_1^{(i)}, x_2^{(i)}))
\end{split}
\end{equation}
where $\mathbf{p}(x_1^{(i)}, x_2^{(i)})$ is the probability of the patches being dissimilar as output by the softmax function. Note that in the loss, we set class weight for the dissimilar class $\gamma$ as 0.2 to penalize incorrectly classified dissimilar pairs less than incorrectly classified similar pairs. This further serves our objective at the anomaly detection phase to have low false positive rates at high true positive rates (where anomalies are denoted positive class). For training, the objective is combined with the standard backpropagation algorithm with the Adam optimizer \cite{kingma_adam:_2014}, saving the best network weights by testing on the validation set (a set of held-out training examples) periodically. The gradient is additive across the twin tails due to tied weights. We use a batch size of 128 with an initial learning rate of 0.001 and train for a maximum of 500 iterations. Xavier-Glorot weight initialization \cite{glorot_understanding_2010} sampling from a normal distribution is used in tandem with ReLU activations in all layers. One important point to note is that, rather than save the network weights that maximize validation accuracy or minimize validation loss, we save that which maximizes validation area under the receiver operating characteristic curve (AUC) for false positive rates up to 0.3. This ROC curve is obtained by plotting true positive rate as a function of false positive rate, where the dissimilar class is denoted positive. By maximizing this AUC, the network that \textit{orders distances in a way that achieves high true positive rate at low false positive rates} is preferred, the behavior we would like to see when it comes time for the anomaly detection phase. We use label smoothing regularization \cite{szegedy_rethinking_2016} set to 0.1 to aid generalization. We find that adding label smoothing regularization is helpful for two reasons. The first is that the video patch pairing process has to in a sense guess what a \textit{future learned function} should call similar and different in order to achieve good performance on anomaly detection, so it produces a dataset with noisy labels. The second arises from the observation that minimizing the cross entropy is equivalent to maximizing the log-likelihood of the correct label, which makes the network try to increase the logit corresponding to the correct label and make it much larger than the other logits, causing it to overfit to the training data and become too confident about its predictions. Label smoothing helps with both of these by making the network less confident about its predictions. We also use dropout \cite{srivastava_dropout:_nodate} of 0.3 on the activations of the second to last fully connected layer (fc1).

\subsection{Exemplar learning and anomaly detection on target dataset}
Detecting anomalies on a target dataset involves two stages: exemplar
model building using the train partition of the dataset and anomaly
detection on the test partition.  Both stages use the previously
trained Siamese network to measure distance between video
patches. This is done by simply treating the softmax of the logit
value that corresponds to the video patches being different as a
measure of distance between the patches. Because the softmax output 
can also be interpreted as a probability, the distance measured can 
also be interpreted as the \textit{probability of patches being 
different}.
We emphasize that the training of the Siamese network is independent
of the exemplar model building and anomaly detection stages.  The
Siamese network is trained on a different set of source datasets than
the target video anomaly detection dataset.

\textbf{Exemplar learning on train partition of target dataset:} 
Since videos contain a large amount of temporal redundancies, we use the exemplar learning approach of \cite{jones_exemplar_2016} to build a model of normal activity in the target dataset.  The exemplar model consists of sets of region-specific exemplar video patches from the videos in the train partition using a sliding spatio-temporal window with spatial stride ($H/2$, $W/2$) and temporal stride of 1. The point of exemplar learning is to represent the set of all video patches in the train partition using a smaller set of unique, representative video patches.  The feature vector learned by the Siamese network is used to represent a video patch and the distance function learned by the Siamese network measures the distance between two feature vectors.  A video patch is added to the exemplar set for a particular spatial region if its distance to the nearest exemplar for that region is above a threshold, which we set to 0.3 for all experiments.  Figure \ref{fig:overview_2} illustrates a subset of exemplar video patches extracted from one region of the camera's field of view in the UCSD Ped1 dataset by our CNN. One big advantage of the exemplar learning approach is that updating the exemplar set in a streaming fashion is possible. This makes the approach scalable and adaptable to environmental changes over time.

\textbf{Anomaly detection on test partition of target dataset}: At test time, overlapping patches with spatial stride ($H/2$, $W/2$) and temporal stride of 1 are extracted from the test partition and distances to nearest exemplars produce anomaly scores (see Figure \ref{fig:overview_2}). In both the exemplar learning and anomaly scoring phases, we achieve additional speedup by ignoring video patches that contain little or no motion. Specifically, a video patch is ignored if under $20\%$ of its pixels across the channel dimension do not satisfy a threshold on flow magnitude or a threshold on the raw pixel value difference between the current and the previous frame. Furthermore, the brute-force nearest neighbor search used in the experiments could be replaced by a fast approximate nearest neighbors algorithm \cite{muja_flann_2009} for further speed-up.  Anomaly scores are stored and aggregated in a pixel map and the final anomaly score of a pixel is simply the mean of all anomaly scores it received as part of patches it participated in (due to overlap of patches in space and time). The anomaly detection is region-specific, so a patch is only compared to exemplars extracted from the same region.


\section{Experiments}
\label{sect:exp}
\subsection{Experimental setup - Datasets and evaluation measures}
\label{subsect:exp-setup}

\textbf{Datasets:} We perform experiments on 3 benchmark datasets: UCSD Ped1, UCSD Ped2 \cite{mahadevan_anomaly_2010} and CUHK Avenue \cite{lu_abnormal_2013}. Each of these datasets includes pre-defined train and test partitions from a single static camera where train partitions contain sequences of normal activity only and test partitions contain sequences with both normal and anomalous activity, and with spatial anomaly annotations per frame.

\textbf{Evaluation measures:}  To compare against other works we use the widely-used \textbf{frame-level} and \textbf{pixel-level} area under the curve (AUC) and equal error rate (EER) criteria proposed in \cite{mahadevan_anomaly_2010}.

In addition, we report performance using two new criteria presented in \cite{ramachandra_street_2019}, which are more representative of real-world performance as argued in that paper. The first is a \textbf{region-based criterion}:
A true positive occurs if a ground truth annotated region has a minimum intersection over union (IOU) of 0.1 with a detection region.  Detected regions are formed as connected components of detected pixels.  The total number of positives is correspondingly the total number of anomalous regions in the test data. A false positive occurs if a detected region simply does not satisfy the minimum IOU threshold of 0.1 with any ground truth region.  The region-based ROC curve plots the true positive rate (which is the fraction of ground truth anomalous {\em regions} detected) versus the false positive rate per frame. The second is a \textbf{track-based criterion}: A true positive occurs if at least 10\% of the frames comprising a ground truth anomaly's track satisfy the region-based criterion.  The total number of positives is the number of ground truth annotated tracks in the test data. False positives are counted identically to the region-based criterion.  The track-based ROC curve plots the true positive rate (which is the fraction of ground truth anomalous {\em tracks} detected) versus the false positive rate per frame. AUCs for both criteria are calculated for false positive rates from 0.0 up to 1.0.  Because the track-based criterion requires ground truth annotations to have a track ID, we relabeled the Ped1, Ped2, and Avenue test sets with bounding boxes that include a track ID.  These new labels will be made publicly available.
Old labels are used for the frame and pixel-level criteria.

\subsection{Comparison against state of the art}
\label{subsect:sota}
\begin{table}
\centering
\resizebox{\linewidth}{!}{%
\begin{tabular}{ll}
\hline
\textbf{Target dataset}                                            & \textbf{Source datasets}                                                                                                                            \\ \hline
UCSD Ped1                                                          & \begin{tabular}[c]{@{}l@{}}Shanghai Tech camera 06 (quarterscale), \\ Shanghai Tech camera 10 (quarterscale), \\ UCSD Ped2 (halfscale, rotated at 45 degrees), \\ CUHK Avenue (quarterscale)\end{tabular} \\ \hline
UCSD Ped2                                                          & UCSD Ped1, CUHK Avenue (halfscale)                                                                                                                       \\ \hline
\begin{tabular}[c]{@{}l@{}}CUHK Avenue \\ (halfscale)\end{tabular} & UCSD Ped1, UCSD Ped2                                                                                                                                \\ \hline
\end{tabular}%
}
\vspace{1pt}
\caption{Source dataset configuration for each target dataset.}
\vspace{-10pt}
\label{tab:dataset-conf}
\end{table}

\begin{table}[h]
\centering
\resizebox{\linewidth}{!}{%
\begin{tabular}{|l|lll||lll|}
  \hline
  {\bf Method} & \multicolumn{3}{c||}{\bf track AUC} & \multicolumn{3}{c|}{\bf region AUC} \\
  & {\bf Ped1} & {\bf Ped2} & {\bf Avenue} & {\bf Ped1} & {\bf Ped2} & {\bf Avenue}\\ \hline
\cite{ramachandra_street_2019} (FG masks) & 84.6\% & 80.5\% & \textbf{80.9\%} & 46.6\% & 62.5\% & 35.8\% \\ \hline
\cite{ramachandra_street_2019} (Flow) & 86.5\% & 83.2\% & 78.4\% & 48.3\% & 55.0\% & 27.3\%\\ \hline
\textbf{Ours} & \textbf{90.0\%} & \textbf{89.3\%} & 78.6\% & \textbf{59.2\%} & \textbf{74.0\%} & \textbf{41.2\%} \\ \hline
\end{tabular}%
}
\vspace{1pt}
\caption{Track and region-based criteria, area under the curve for false positive rates up to 1.0.}
\label{tab:track-region-aucs}
\end{table}

To evaluate our approach, we compare against results reported on the traditional evaluation measures by papers in the recent literature. For each of our experiments, a new CNN was trained using only datasets other than the target dataset to curate the training data for the Siamese network (see Table \ref{tab:dataset-conf}), but each newly trained network used the same aforementioned regularization parameters. A simple heuristic was used to choose which source datasets should be used for a given target dataset - those datasets in which the scale of objects roughly match that in the target dataset for a $H \times W$ image patch.  In future work, we plan to use more labeled videos to train a single Siamese network that works well across many different target datasets.

Table \ref{tab:auc-table} presents frame and pixel-level AUC measures on the UCSD Ped1, UCSD Ped2 and CUHK Avenue datasets. Our approach sets new state of the art on UCSD Ped2 pixel-level AUC by around $4\%$ as well as on CUHK Avenue frame EER by around $6\%$. Upon visualizing the detections, we find that our approach finds it particularly difficult to detect anomalies at very small scales that exist in the UCSD Ped1 test set.  Also, our method, like most others in Table \ref{tab:auc-table}, is unable to detect loitering anomalies present in the CUHK Avenue dataset.  This is mainly due to our use of a ``motion check'' that ignores video patches with little or no motion for efficiency reasons.  This could be replaced by a more sophisticated background model that is slower to absorb stationary objects.


Further, we report AUC for false positive rates up to 1.0 for the track and region based criteria in Table \ref{tab:track-region-aucs}. We reimplemented the work of \cite{ramachandra_street_2019} for these results. Clearly, our approach surpasses that of \cite{ramachandra_street_2019}, meaning we detect more anomalous events (tracks and regions) while also producing fewer false positives per frame overall.

\begin{figure}[ht]
\centering
\includegraphics[width=\linewidth]{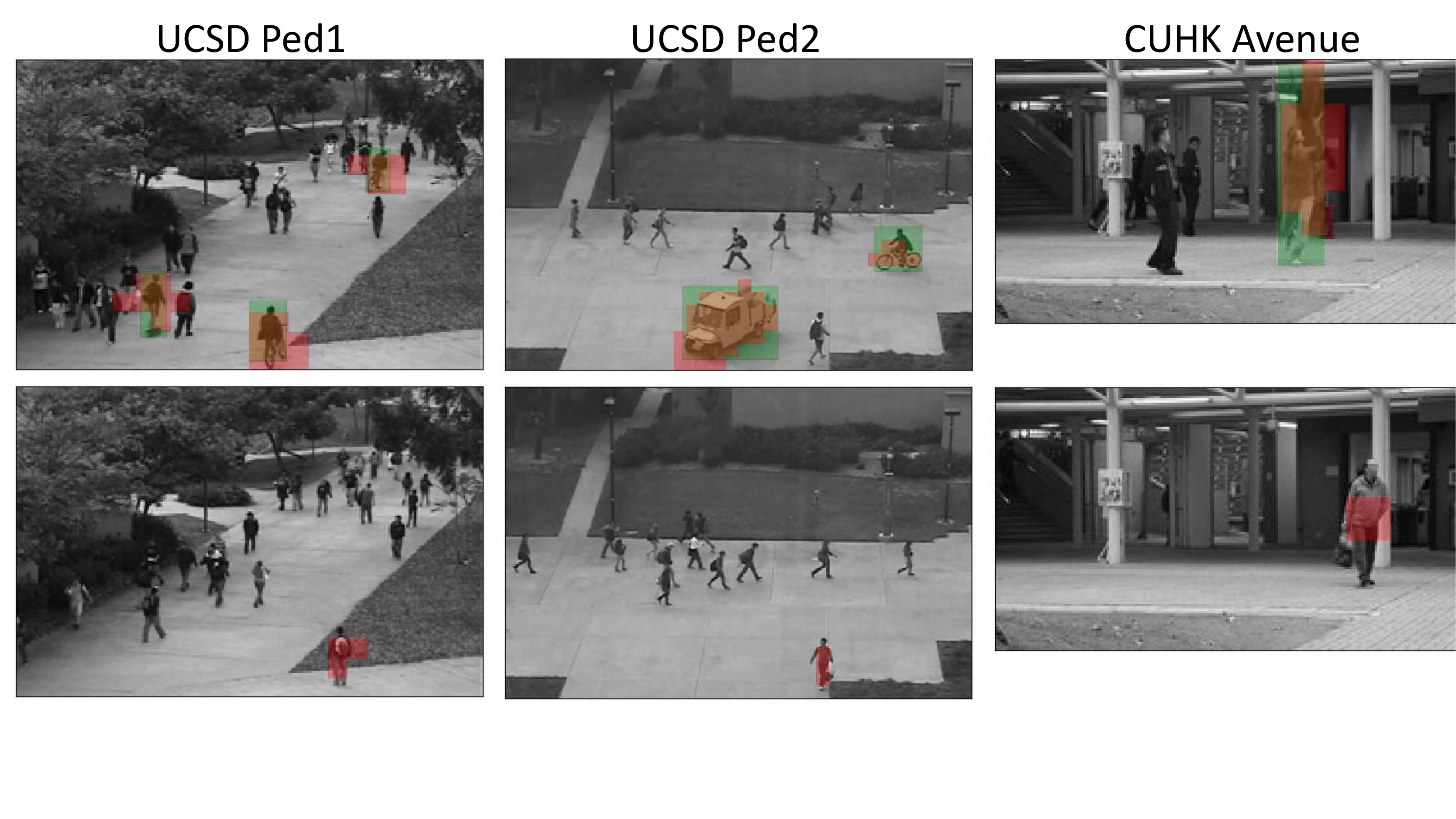}
\vspace{-28pt}
\caption{Examples of true positives (first row) and false positives (second row) from our detector on all 3 datasets. Green bounding box annotations denote ground truth anomalies and red regions our model's detections (intersections are orange-ish).}
\vspace{-10pt}
\label{fig:tp_fp}
\end{figure}

\begin{figure}[ht]
\centering
\includegraphics[width=\linewidth]{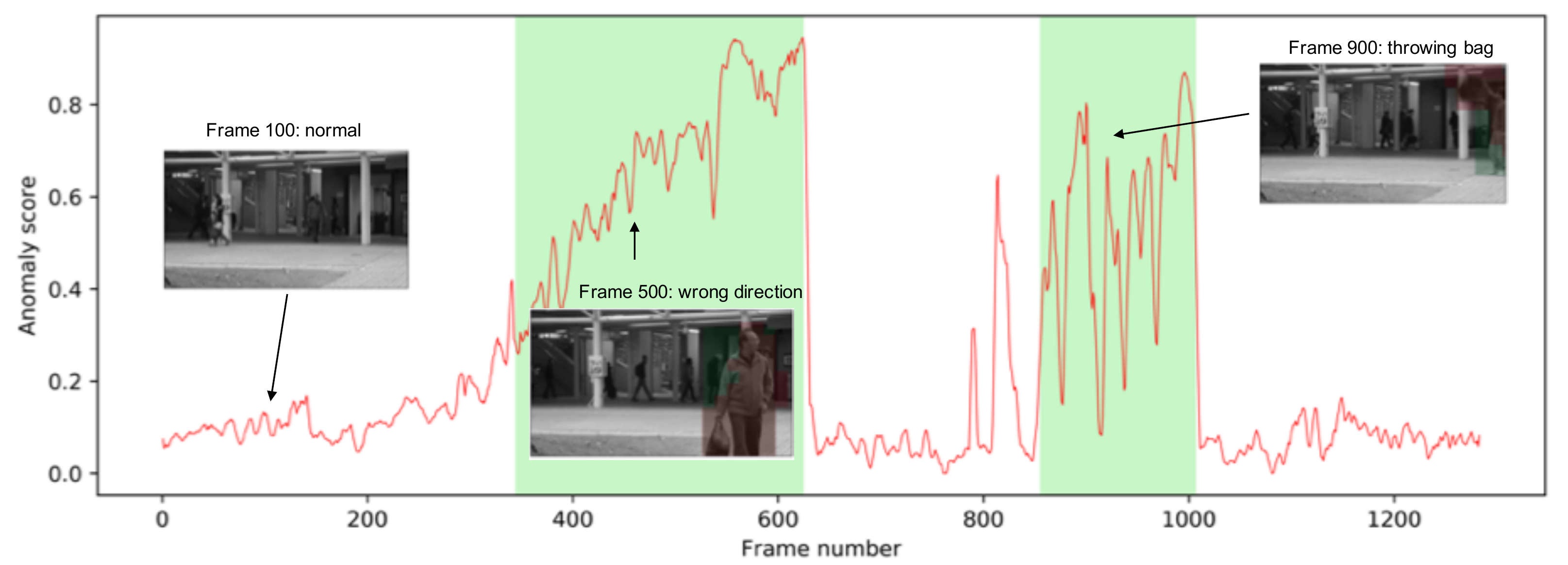}
\caption{Anomaly score as a function of frame number for CUHK Avenue Test sequence number 6. Green shading on the plot denotes ground truth anomalous frames.}
\vspace{-10pt}
\label{fig:pfe_avenue}
\end{figure}

These ROC curves and AUC measures do not completely capture the behavior of video anomaly detection approaches. In \cite{lobo_auc:_2008}, the authors present an excellent analysis of the problems with an evaluation measure such as AUC. Thus, we present a set of qualitative results here. Figure \ref{fig:tp_fp} shows some detection results at a fixed anomaly score threshold. We notice that the {\em quality of false positives} in our approach is high, and often we are able to attribute reasons for these errors. For example, the false positive shown in the figure for UCSD Ped1 dataset is due to the fact that a person is never seen walking across the grass in this specific manner in the train partition. A similar argument explains the false positives shown for the other two datasets as well. This could either indicate that the train partition is incomplete, or highlight the subjectivity involved in ground truth annotation processes. Figure \ref{fig:pfe_avenue} illustrates how anomaly score per frame, computed as the maximum of anomaly scores of pixels in the frame, varies for one test sequence of CUHK Avenue. The high variance in anomaly scores during the ``bag throwing'' anomaly even indicates how this event might intersperse normal and anomalous frames, seeming normal when the bag leaves the camera frame and vice versa.

\subsection{Ablation study on source datasets used}
\label{subsect:understanding}

\begin{table}[h]
\centering
\resizebox{\linewidth}{!}{%
\begin{tabular}{|ccc||cc|}
  \hline
  \multicolumn{3}{|c||}{\bf Source datasets} & \multicolumn{2}{c|}{\bf Target = Ped2} \\
  {\bf Ped1} & {\bf Avenue} & {\bf ShanghaiTech} & {\bf Frame AUC} & {\bf Pixel AUC}\\ \hline
  Y & & & 90.9\% & 89.4\% \\ \hline
   & Y & & 90.4\% & 88.7\% \\ \hline
   & & Y & 93.7\% & 93.0\% \\ \hline
  Y & Y & & 94.0\% & 93.0\% \\ \hline
   & Y & Y & 91.8\% & 91.0\% \\ \hline
  Y &  & Y & 91.7\% & 90.7\% \\ \hline
  Y & Y & Y & 93.0\% & 91.9\% \\ \hline
\end{tabular}%
}
\vspace{1pt}
\caption{Ablation study on the choice of source datasets for a particular target dataset. `Y` denotes that the dataset was used in the source pool.}
\label{tab:source-ablation}
\end{table}

We perform an ablation study to understand the effect of picking source datasets for a particular target dataset. Since it is prohibitive to perform a complete ablation study, for this study we set the target to be UCSD Ped2 and vary all non-empty subsets of source datasets from the set \{UCSD Ped1, CUHK Avenue, ShanghaiTech (cameras 06 and 10)\}, training only once. The results presented in Table \ref{tab:source-ablation} show that while there is some sensitivity to the choice of source datasets, on both the frame and pixel level measures, we see a variation of $<5\%$. This variation is from a combination of variation due to stochasticity during training (batching, random initialization, dropout) and choice of source datasets. 

\section{Conclusion}
\label{sect:concl}
We have presented a novel approach to video anomaly detection that
introduces a new way to use a deep network for this problem.  We
substitute the problem of classifying a video patch as anomalous or
not for the problem of estimating a distance between two video 
patches, for which we can generate plenty of labeled training data.  
The learned
distance function (which also learns a feature vector to represent a
video patch) can then be used in a straightforward video anomaly
detection method that measures the distance from each testing video
patch to the nearest exemplar video patch for that region. We have
shown that our approach either surpasses or performs comparably to the previous state of the art without any training of the Siamese network on data from the target dataset.  Our approach also possesses some favorable 
properties
in being a plug-and-play method (learned distance function can be used
out-of-the-box on target dataset), and in being scalable and
resistant to environmental changes (updation of the exemplar set is
easy).

{\small
\bibliographystyle{ieee}
\bibliography{VADSiamese,extras}

\begin{thebibliography}{10}\itemsep=-1pt

\bibitem{adam_robust_2008}
A.~Adam, E.~Rivlin, I.~Shimshoni, and D.~Reinitz.
\newblock Robust {Real}-{Time} {Unusual} {Event} {Detection} using {Multiple}
  {Fixed}-{Location} {Monitors}.
\newblock {\em IEEE Transactions on Pattern Analysis and Machine Intelligence},
  30(3):555--560, Mar. 2008.

\bibitem{antic_video_2011}
B.~Antic and B.~Ommer.
\newblock Video parsing for abnormality detection.
\newblock In {\em 2011 {International} {Conference} on {Computer} {Vision}},
  pages 2415--2422, Barcelona, Spain, Nov. 2011. IEEE.

\bibitem{antic_spatio-temporal_2015}
B.~Anti{\'c} and B.~Ommer.
\newblock Spatio-temporal {Video} {Parsing} for {Abnormality} {Detection}.
\newblock {\em arXiv preprint arXiv:1502.06235}, 2015.

\bibitem{bromley_signature_1994}
J.~Bromley, I.~Guyon, Y.~LeCun, E.~S{\"a}ckinger, and R.~Shah.
\newblock Signature verification using a siamese time delay neural network.
\newblock In {\em Advances in neural information processing systems}, pages
  737--744, 1994.

\bibitem{chong_abnormal_2017}
Y.~S. Chong and Y.~H. Tay.
\newblock Abnormal {Event} {Detection} in {Videos} using {Spatiotemporal}
  {Autoencoder}.
\newblock {\em arXiv:1701.01546 [cs]}, Jan. 2017.
\newblock arXiv: 1701.01546.

\bibitem{cong_abnormal_2013}
Y.~Cong, J.~Yuan, and J.~Liu.
\newblock Abnormal event detection in crowded scenes using sparse
  representation.
\newblock {\em Pattern Recognition}, 46(7):1851--1864, July 2013.

\bibitem{leibe_discriminative_2016}
A.~Del~Giorno, J.~A. Bagnell, and M.~Hebert.
\newblock A {Discriminative} {Framework} for {Anomaly} {Detection} in {Large}
  {Videos}.
\newblock In {\em European Conference on Computer Vision (ECCV)}, pages
  334--349. 2016.

\bibitem{dollar2005behavior}
P.~Doll{\'a}r, V.~Rabaud, G.~Cottrell, and S.~Belongie.
\newblock Behavior recognition via sparse spatio-temporal features.
\newblock VS-PETS Beijing, China, 2005.

\bibitem{feng_learning_2017}
Y.~Feng, Y.~Yuan, and X.~Lu.
\newblock Learning deep event models for crowd anomaly detection.
\newblock {\em Neurocomputing}, 219:548--556, Jan. 2017.

\bibitem{glorot_understanding_2010}
X.~Glorot and Y.~Bengio.
\newblock Understanding the difficulty of training deep feedforward neural
  networks.
\newblock In {\em Proceedings of Machine Learning Research (PMLR)}, pages
  249--256, Mar. 2010.

\bibitem{han_matchnet_2015}
X.~Han, T.~Leung, Y.~Jia, R.~Sukthankar, and A.~C. Berg.
\newblock Matchnet: Unifying feature and metric learning for patch-based
  matching.
\newblock In {\em IEEE Conference on Computer Vision and Pattern Recognition
  (CVPR)}, pages 3279--3286, 2015.

\bibitem{hasan_learning_2016}
M.~Hasan, J.~Choi, J.~Neumann, A.~K. Roy-Chowdhury, and L.~S. Davis.
\newblock Learning {Temporal} {Regularity} in {Video} {Sequences}.
\newblock In {\em {IEEE} {Conference} on {Computer} {Vision} and {Pattern}
  {Recognition} ({CVPR})}, pages 733--742, Las Vegas, NV, USA, June 2016.

\bibitem{hinami_joint_2017}
R.~Hinami, T.~Mei, and S.~Satoh.
\newblock Joint {Detection} and {Recounting} of {Abnormal} {Events} by
  {Learning} {Deep} {Generic} {Knowledge}.
\newblock In {\em {IEEE} {International} {Conference} on {Computer} {Vision}
  ({ICCV})}, pages 3639--3647, Venice, Oct. 2017.

\bibitem{ionescu_object-centric_2019}
R.~T. Ionescu, F.~S. Khan, M.-I. Georgescu, and L.~Shao.
\newblock Object-centric auto-encoders and dummy anomalies for abnormal event
  detection in video.
\newblock In {\em IEEE {Conference} on {Computer} {Vision} and {Pattern}
  {Recognition} (CVPR)}, pages 7842--7851, 2019.

\bibitem{ionescu_unmasking_2017}
R.~T. Ionescu, S.~Smeureanu, B.~Alexe, and M.~Popescu.
\newblock Unmasking the {Abnormal} {Events} in {Video}.
\newblock In {\em 2017 {IEEE} {International} {Conference} on {Computer}
  {Vision} ({ICCV})}, pages 2914--2922, Venice, Oct. 2017. IEEE.

\bibitem{ionescu_detecting_2019}
R.~T. Ionescu, S.~Smeureanu, M.~Popescu, and B.~Alexe.
\newblock Detecting {Abnormal} {Events} in {Video} {Using} {Narrowed}
  {Normality} {Clusters}.
\newblock In {\em {IEEE} {Winter} {Conference} on {Applications} of {Computer}
  {Vision} ({WACV})}, pages 1951--1960, Jan. 2019.

\bibitem{jones_exemplar_2016}
M.~Jones, D.~Nikovski, M.~Imamura, and T.~Hirata.
\newblock Exemplar learning for extremely efficient anomaly detection in
  real-valued time series.
\newblock {\em Data Mining and Knowledge Discovery (DMKD)}, 30(6):1427--1454,
  2016.

\bibitem{kingma_adam:_2014}
D.~Kingma and J.~Ba.
\newblock Adam: {A} method for stochastic optimization.
\newblock {\em arXiv preprint arXiv:1412.6980}, 2014.

\bibitem{leyva_video_2017}
R.~Leyva, V.~Sanchez, and C.-T. Li.
\newblock Video {Anomaly} {Detection} {With} {Compact} {Feature} {Sets} for
  {Online} {Performance}.
\newblock {\em IEEE Transactions on Image Processing}, 26(7):3463--3478, July
  2017.

\bibitem{liu_beyond_nodate}
C.~Liu.
\newblock Beyond {Pixels}: {Exploring} {New} {Representations} and
  {Applications} for {Motion} {Analysis}.
\newblock {\em MIT PhD Thesis}, 2009.

\bibitem{liu_future_2018}
W.~Liu, W.~Luo, D.~Lian, and S.~Gao.
\newblock Future frame prediction for anomaly detection{\textendash}a new
  baseline.
\newblock In {\em {IEEE} {Conference} on {Computer} {Vision} and {Pattern}
  {Recognition} (CVPR)}, pages 6536--6545, 2018.

\bibitem{liu_classier_2018}
Y.~Liu, C.-L. Li, and B.~P{\'o}czos.
\newblock Classifier {Two}-{Sample} {Test} for {Video} {Anomaly} {Detections}.
\newblock page~12, 2018.

\bibitem{lobo_auc:_2008}
J.~M. Lobo, A.~Jim{\'e}nez-Valverde, and R.~Real.
\newblock {AUC}: a misleading measure of the performance of predictive
  distribution models.
\newblock {\em Global Ecology and Biogeography}, 17(2):145--151, 2008.

\bibitem{lu_abnormal_2013}
C.~Lu, J.~Shi, and J.~Jia.
\newblock Abnormal {Event} {Detection} at 150 {FPS} in {MATLAB}.
\newblock In {\em IEEE International Conference on Computer Vision (ICCV)},
  pages 2720--2727, Sydney, Australia, Dec. 2013.

\bibitem{luo_revisit_2017}
W.~Luo, W.~Liu, and S.~Gao.
\newblock A {Revisit} of {Sparse} {Coding} {Based} {Anomaly} {Detection} in
  {Stacked} {RNN} {Framework}.
\newblock In {\em {IEEE} {International} {Conference} on {Computer} {Vision}
  ({ICCV})}, pages 341--349, Venice, Oct. 2017. IEEE.

\bibitem{mahadevan_anomaly_2010}
V.~Mahadevan, W.~Li, V.~Bhalodia, and N.~Vasconcelos.
\newblock Anomaly detection in crowded scenes.
\newblock In {\em 2010 {IEEE} {Computer} {Society} {Conference} on {Computer}
  {Vision} and {Pattern} {Recognition}}, pages 1975--1981, June 2010.

\bibitem{mehran_abnormal_2009}
R.~Mehran, A.~Oyama, and M.~Shah.
\newblock Abnormal crowd behavior detection using social force model.
\newblock In {\em IEEE Conference on Computer Vision and Pattern Recognition
  (CVPR)}, pages 935--942. IEEE, 2009.

\bibitem{muja_flann_2009}
M.~Muja and D.~G. Lowe.
\newblock Fast approximate nearest neighbors with automatic algorithm
  configuration.
\newblock In {\em International Conference on Computer Vision Theory and
  Application VISSAPP'09)}, pages 331--340. INSTICC Press, 2009.

\bibitem{ramachandra_street_2019}
B.~Ramachandra and M.~Jones.
\newblock Street {Scene}: {A} new dataset and evaluation protocol for video
  anomaly detection.
\newblock In {\em {IEEE} {Winter} {Conference} on {Applications} of {Computer}
  {Vision} ({WACV})}, 2020.

\bibitem{ravanbakhsh_plug-and-play_2018}
M.~Ravanbakhsh, M.~Nabi, H.~Mousavi, E.~Sangineto, and N.~Sebe.
\newblock Plug-and-{Play} {CNN} for {Crowd} {Motion} {Analysis}: {An}
  {Application} in {Abnormal} {Event} {Detection}.
\newblock In {\em {IEEE} {Winter} {Conference} on {Applications} of {Computer}
  {Vision} ({WACV})}, pages 1689--1698, Lake Tahoe, NV, Mar. 2018.

\bibitem{ravanbakhsh_abnormal_2017}
M.~Ravanbakhsh, M.~Nabi, E.~Sangineto, L.~Marcenaro, C.~Regazzoni, and N.~Sebe.
\newblock Abnormal event detection in videos using generative adversarial nets.
\newblock In {\em 2017 {IEEE} {International} {Conference} on {Image}
  {Processing} ({ICIP})}, pages 1577--1581, Sept. 2017.

\bibitem{saligrama_video_2012}
V.~Saligrama and Z.~Chen.
\newblock Video anomaly detection based on local statistical aggregates.
\newblock In {\em IEEE Conference on Computer Vision and Pattern Recognition
  (CVPR)}, pages 2112--2119. IEEE, 2012.

\bibitem{smeureanu_deep_2017}
S.~Smeureanu, R.~T. Ionescu, M.~Popescu, and B.~Alexe.
\newblock Deep {Appearance} {Features} for {Abnormal} {Behavior} {Detection} in
  {Video}.
\newblock In S.~Battiato, G.~Gallo, R.~Schettini, and F.~Stanco, editors, {\em
  International Conference on Image {Analysis} and {Processing} (ICIAP)}, pages
  779--789. Springer International Publishing, Cham, 2017.

\bibitem{srivastava_dropout:_nodate}
N.~Srivastava, G.~Hinton, A.~Krizhevsky, I.~Sutskever, and R.~Salakhutdinov.
\newblock Dropout: {A} {Simple} {Way} to {Prevent} {Neural} {Networks} from
  {Overfitting}.
\newblock {\em Journal of Machine Learning Research}, 2014.

\bibitem{sultani_real-world_2018}
W.~Sultani, C.~Chen, and M.~Shah.
\newblock Real-{World} {Anomaly} {Detection} in {Surveillance} {Videos}.
\newblock In {\em 2018 {IEEE}/{CVF} {Conference} on {Computer} {Vision} and
  {Pattern} {Recognition}}, pages 6479--6488, Salt Lake City, UT, June 2018.
  IEEE.

\bibitem{szegedy_rethinking_2016}
C.~Szegedy, V.~Vanhoucke, S.~Ioffe, J.~Shlens, and Z.~Wojna.
\newblock Rethinking the {Inception} {Architecture} for {Computer} {Vision}.
\newblock In {\em {IEEE} {Conference} on {Computer} {Vision} and {Pattern}
  {Recognition} ({CVPR})}, pages 2818--2826, Las Vegas, NV, USA, June 2016.
  IEEE.

\bibitem{turchini_convex_2017}
F.~Turchini, L.~Seidenari, and A.~Del~Bimbo.
\newblock Convex {Polytope} {Ensembles} for {Spatio}-{Temporal} {Anomaly}
  {Detection}.
\newblock In S.~Battiato, G.~Gallo, R.~Schettini, and F.~Stanco, editors, {\em
  International Conference on Image {Analysis} and {Processing} (ICIAP)},
  Lecture {Notes} in {Computer} {Science}, pages 174--184. Springer
  International Publishing, 2017.

\bibitem{weixin_li_anomaly_2014}
{Weixin Li}, V.~Mahadevan, and N.~Vasconcelos.
\newblock Anomaly {Detection} and {Localization} in {Crowded} {Scenes}.
\newblock {\em IEEE Transactions on Pattern Analysis and Machine Intelligence},
  36(1):18--32, Jan. 2014.

\bibitem{wu_chaotic_2010}
S.~Wu, B.~E. Moore, and M.~Shah.
\newblock Chaotic invariants of lagrangian particle trajectories for anomaly
  detection in crowded scenes.
\newblock In {\em IEEE Conference on Computer {Vision} and {Pattern}
  {Recognition} ({CVPR})}, pages 2054--2060. IEEE, 2010.

\bibitem{xu_learning_2015}
D.~Xu, E.~Ricci, Y.~Yan, J.~Song, and N.~Sebe.
\newblock Learning deep representations of appearance and motion for anomalous
  event detection.
\newblock {\em arXiv preprint arXiv:1510.01553}, 2015.

\bibitem{zagoruyko_learning_2015}
S.~Zagoruyko and N.~Komodakis.
\newblock Learning to compare image patches via convolutional neural networks.
\newblock In {\em 2015 {IEEE} {Conference} on {Computer} {Vision} and {Pattern}
  {Recognition} ({CVPR})}, pages 4353--4361, Boston, MA, USA, June 2015. IEEE.

\end{thebibliography}
}

\section{Supplemental Material}
\subsection{Understanding the distance function learned}
\begin{figure}[ht]
\centering
\includegraphics[width=\linewidth]{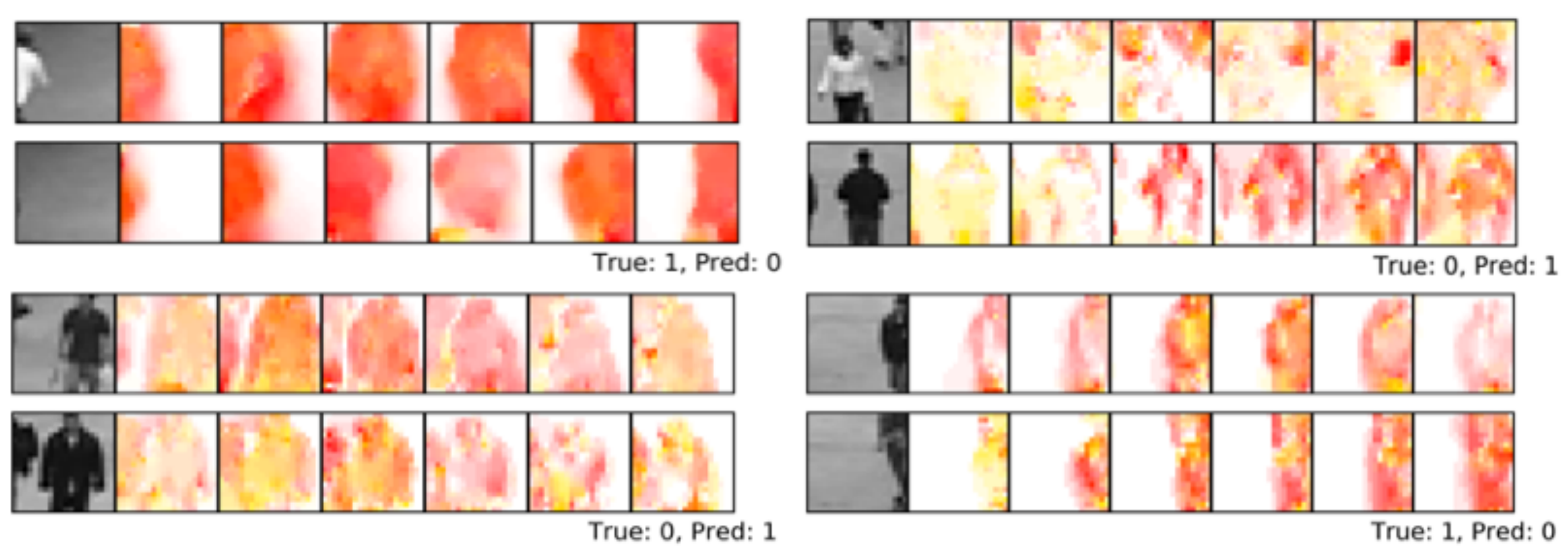}
\caption{Examples of large prediction errors made by our model on UCSD Ped1. Classes 0 and 1 refer to similar and dissimilar pairs respectively. Best viewed in color.}
\label{fig:large_errors}
\end{figure}

We also tried to gain some insight into what properties the distance function learned by the CNN possesses. To this end, we recorded the video patch pairs on which the CNN makes large errors, that is, either classifying similar pairs as dissimilar or vice versa, with high predicted probability. Figure \ref{fig:large_errors} is a visualization of 4 such video patch pairs when the target dataset is UCSD Ped1. Remarkably, the CNN seems to find it hard to correctly classify examples that are conceivably hard for humans. Specifically, the dissimilar pairs that have been misclassified seem to contain a skateboarder moving only slightly faster than a pedestrian would, and the similar pairs that have been misclassified exhibit some distinct differences in their flow fields.

\subsection{Track and region based ROC curves}
\begin{figure}[h]
\centering
\includegraphics[width=\linewidth]{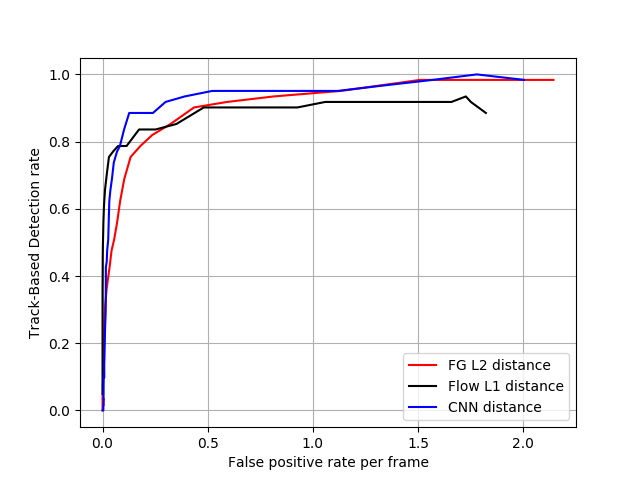}
\caption{Track-based ROC curves on UCSD Ped1.}
\label{fig:ped1_track}
\end{figure}

\begin{figure}[h]
\centering
\includegraphics[width=\linewidth]{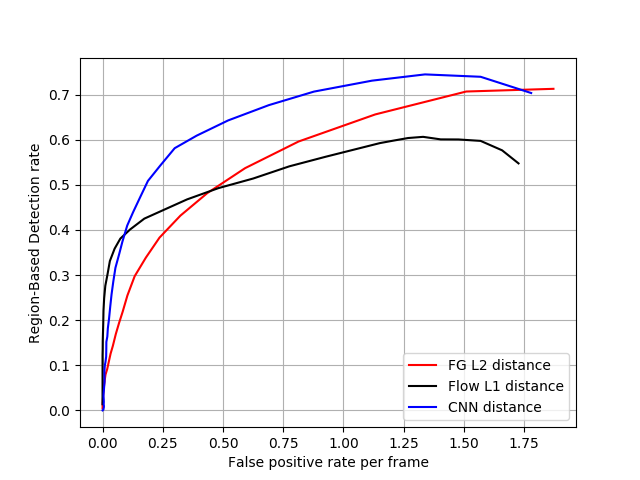}
\caption{Region-based ROC curves on UCSD Ped1.}
\label{fig:ped1_region}
\end{figure}

\begin{figure}[h]
\centering
\includegraphics[width=\linewidth]{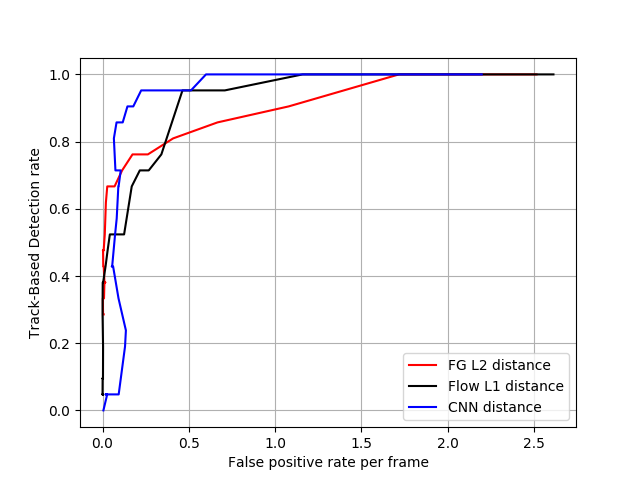}
\caption{Track-based ROC curves on UCSD Ped2.}
\label{fig:ped2_track}
\end{figure}

\begin{figure}[h]
\centering
\includegraphics[width=\linewidth]{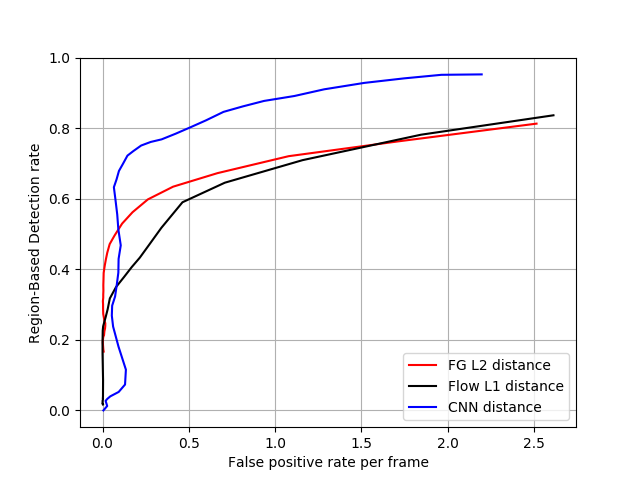}
\caption{Region-based ROC curves on UCSD Ped2.}
\label{fig:ped2_region}
\end{figure}

\begin{figure}[h]
\centering
\includegraphics[width=\linewidth]{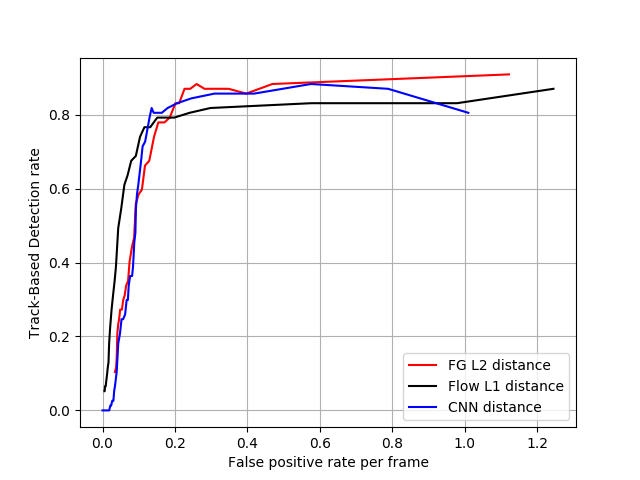}
\caption{Track-based ROC curves on CUHK Avenue.}
\label{fig:avenue_track}
\end{figure}

\begin{figure}[h]
\centering
\includegraphics[width=\linewidth]{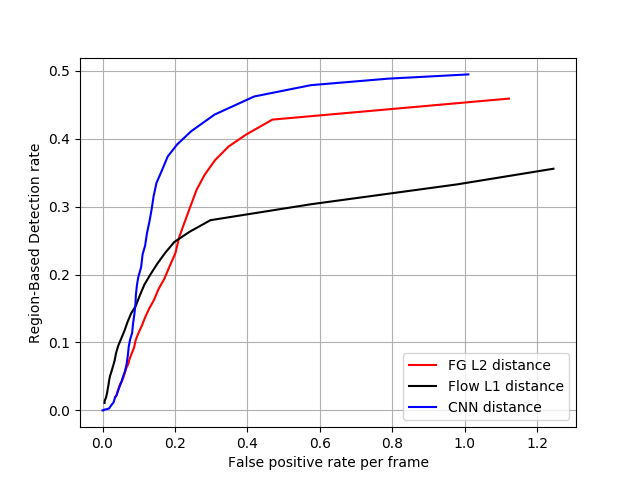}
\caption{Region-based ROC curves on CUHK Avenue.}
\label{fig:avenue_region}
\end{figure}

Figures \ref{fig:ped1_track} through \ref{fig:avenue_region} show the ROC curves for our CNN approach (denoted ``CNN distance'') as well as that of \cite{ramachandra_street_2019}'s FG masks (denoted ``FG L2 distance'') and flow (denoted ``Flow L1 distance'') methods on all 3 datasets. Overall, it appears that our approach of using a learned representation and learned distance function is able to achieve better detection performance, demonstrated by higher true positive rates at low false positive rates.

\subsection{More detection result visualizations}
\begin{figure}[h]
\centering
\includegraphics[width=\linewidth]{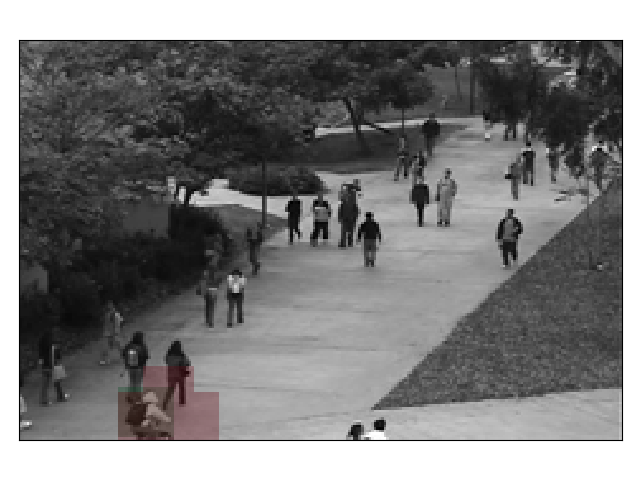}
\caption{True positive in UCSD Ped1 - a biker.}
\label{fig:ped1_tp_1}
\end{figure}

\begin{figure}[h]
\centering
\includegraphics[width=\linewidth]{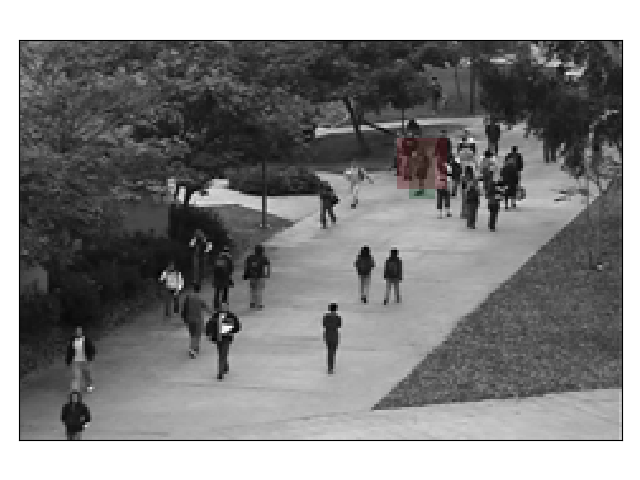}
\caption{True positive in UCSD Ped1 - a skateboarder.}
\label{fig:ped1_tp_2}
\end{figure}

\begin{figure}[h]
\centering
\includegraphics[width=\linewidth]{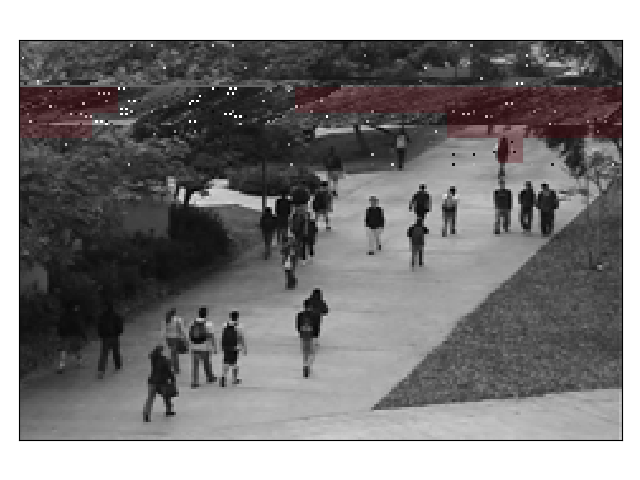}
\caption{False positive in UCSD Ped1 - camera fault.}
\label{fig:ped1_fp_1}
\end{figure}

\begin{figure}[h]
\centering
\includegraphics[width=\linewidth]{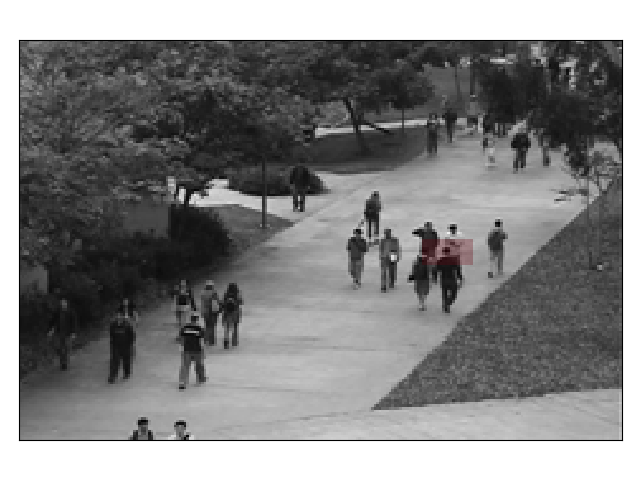}
\caption{False positive in UCSD Ped1 - seemingly random.}
\label{fig:ped1_fp_2}
\end{figure}

\begin{figure}[h]
\centering
\includegraphics[width=\linewidth]{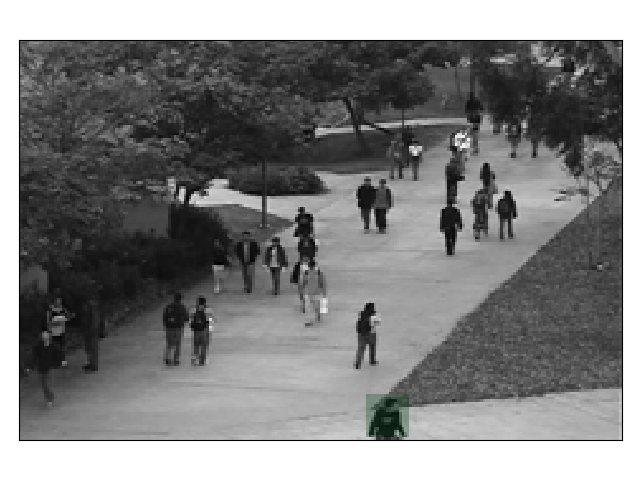}
\caption{False negative in UCSD Ped1 - biker not yet fully in the camera frame.}
\label{fig:ped1_fn_1}
\end{figure}

\begin{figure}[h]
\centering
\includegraphics[width=\linewidth]{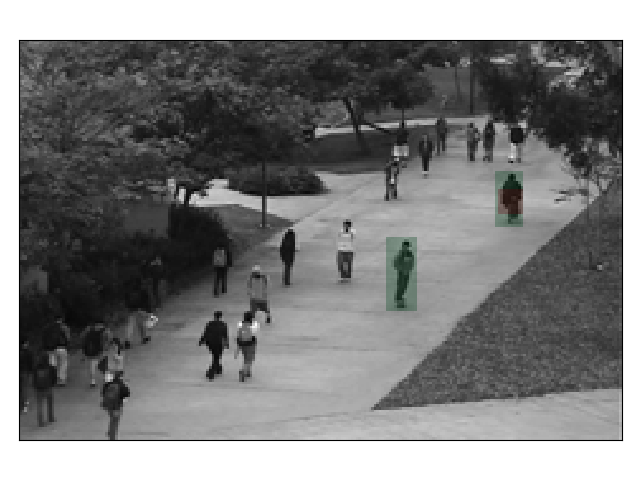}
\caption{False negative in UCSD Ped1 - skateboarder moving slowly.}
\label{fig:ped1_fn_2}
\end{figure}

\begin{figure}[h]
\centering
\includegraphics[width=\linewidth]{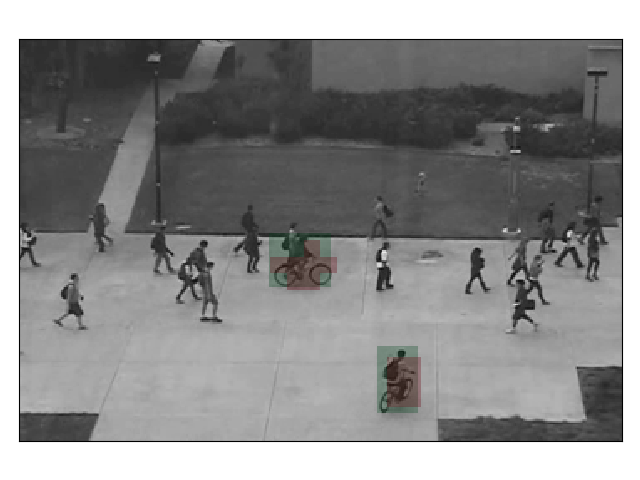}
\caption{True positive in UCSD Ped2 - 2 bikers.}
\label{fig:ped2_tp_1}
\end{figure}

\begin{figure}[ht]
\centering
\includegraphics[width=\linewidth]{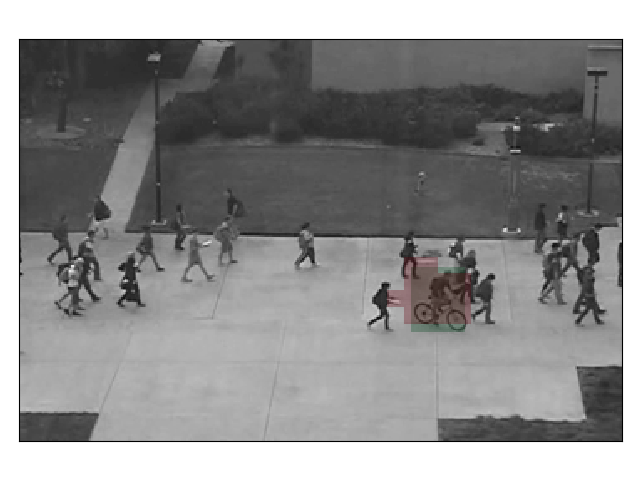}
\caption{True positive in UCSD Ped2 - a biker.}
\label{fig:ped2_tp_2}
\end{figure}

\begin{figure}[h]
\centering
\includegraphics[width=\linewidth]{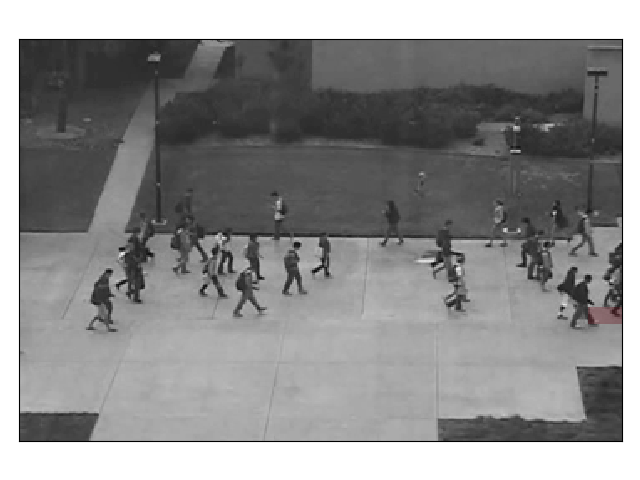}
\caption{False positive in UCSD Ped2 - seemingly random.}
\label{fig:ped2_fp_1}
\end{figure}

\begin{figure}[h]
\centering
\includegraphics[width=\linewidth]{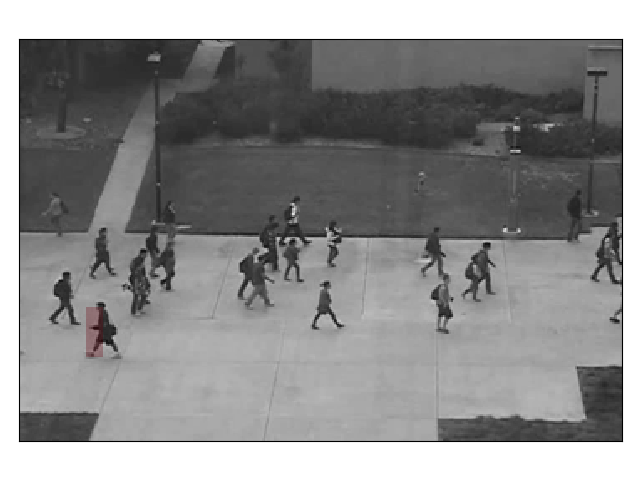}
\caption{False positive in UCSD Ped2 - unusual movement in this region of the camera frame.}
\label{fig:ped2_fp_2}
\end{figure}

\begin{figure}[h]
\centering
\includegraphics[width=\linewidth]{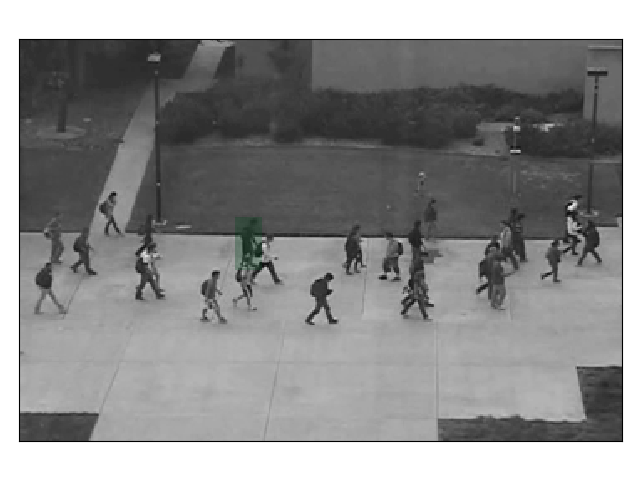}
\caption{False negative in UCSD Ped2 - occluded, slow-moving skateboarder.}
\label{fig:ped2_fn_1}
\end{figure}

\begin{figure}[h]
\centering
\includegraphics[width=\linewidth]{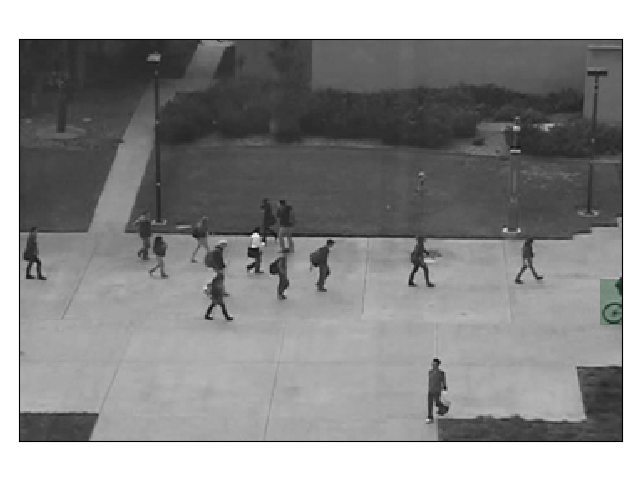}
\caption{False negative in UCSD Ped2 - biker partially left the camera frame.}
\label{fig:ped2_fn_2}
\end{figure}

\begin{figure}[h]
\centering
\includegraphics[width=\linewidth]{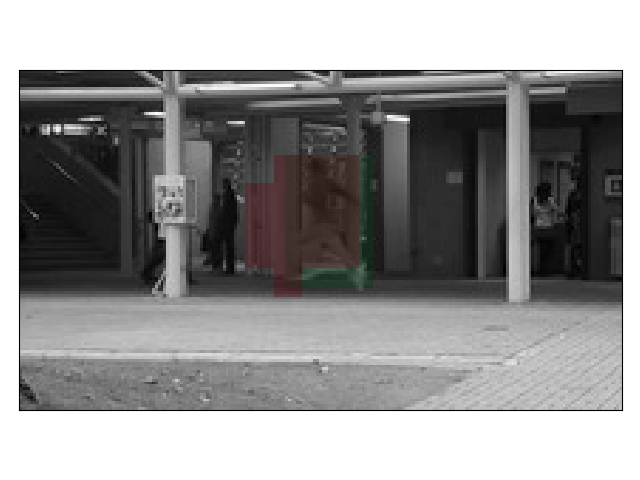}
\caption{True positive in CUHK Avenue - person running.}
\label{fig:avenue_tp_1}
\end{figure}

\begin{figure}[h]
\centering
\includegraphics[width=\linewidth]{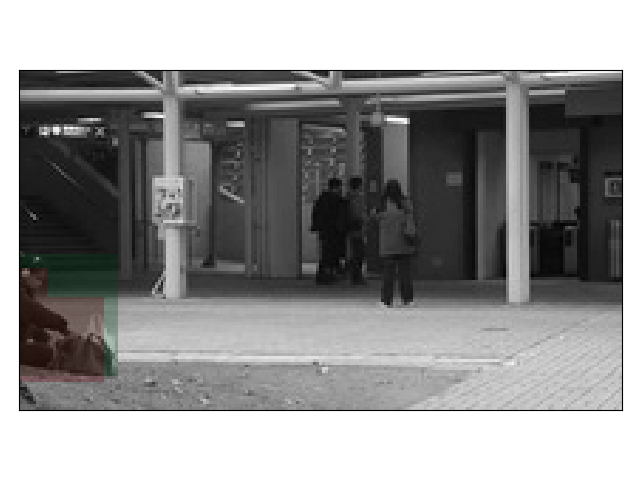}
\caption{True positive in CUHK Avenue - person interacting with a bag on the grass.}
\label{fig:avenue_tp_2}
\end{figure}

\begin{figure}[h]
\centering
\includegraphics[width=\linewidth]{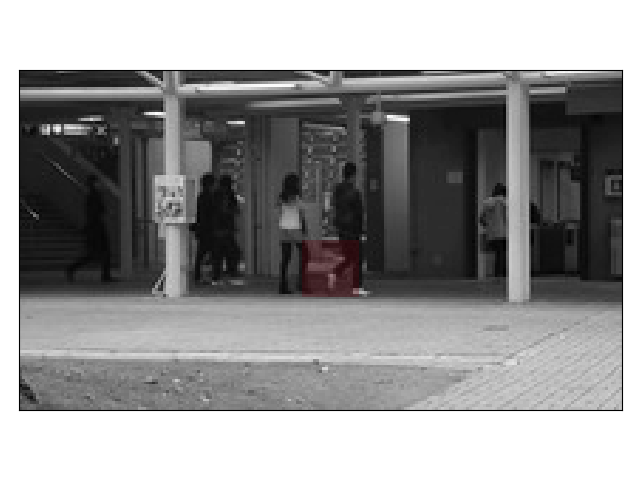}
\caption{False positive in CUHK Avenue - seemingly random.}
\label{fig:avenue_fp_1}
\end{figure}

\begin{figure}[h]
\centering
\includegraphics[width=\linewidth]{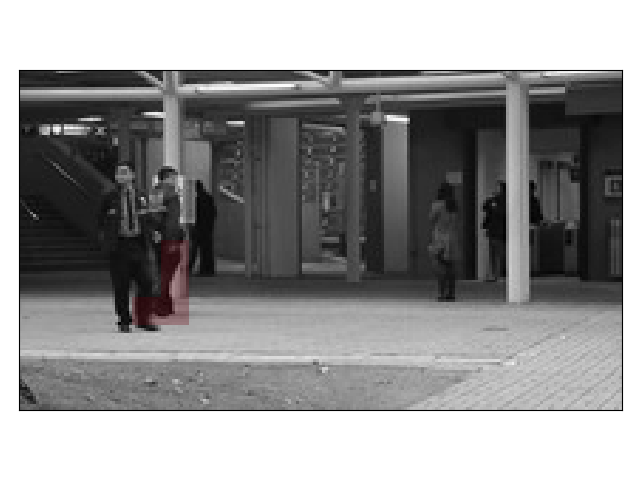}
\caption{False positive in CUHK Avenue - unusual movement in this region of the camera frame.}
\label{fig:avenue_fp_2}
\end{figure}

\begin{figure}[h]
\centering
\includegraphics[width=\linewidth]{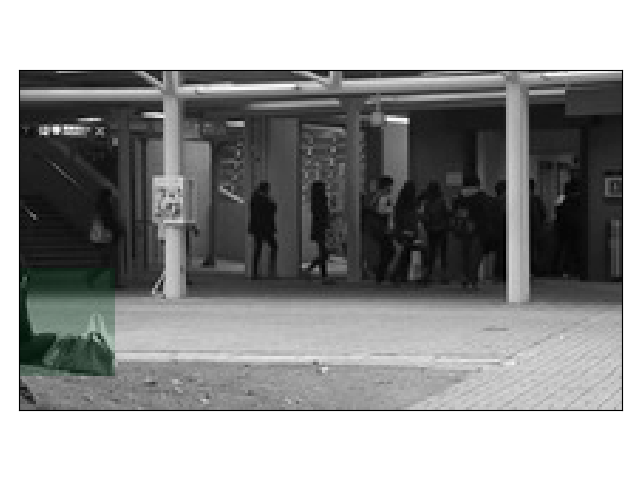}
\caption{False negative in CUHK Avenue - still, unattended bag.}
\label{fig:avenue_fn_1}
\end{figure}

\begin{figure}[h]
\centering
\includegraphics[width=\linewidth]{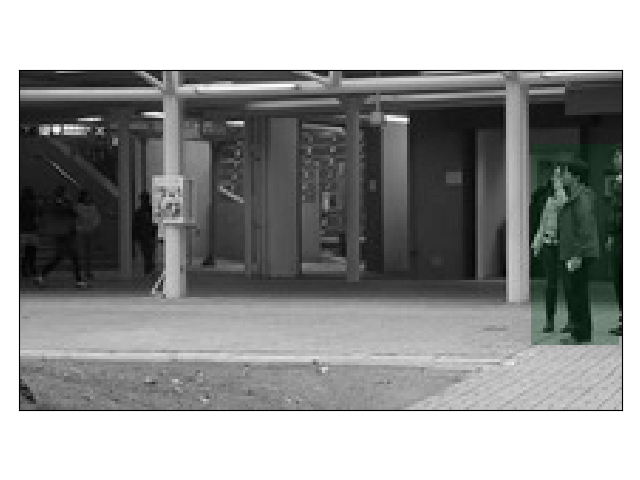}
\caption{False negative in CUHK Avenue - start of an anomalous event that is seemingly normal.}
\label{fig:avenue_fn_2}
\end{figure}

Figures \ref{fig:ped1_tp_1} through \ref{fig:avenue_fn_2} present additional true positive, false positive and false negative detection results from our approach for all 3 datasets. As in the submission document, the green bounding boxes refer to ground truth anomalies and the red regions our detections at a fixed threshold on anomaly scores.


\subsection{More frame-level anomaly score visualizations}
\begin{figure*}
\centering
\includegraphics[width=\linewidth]{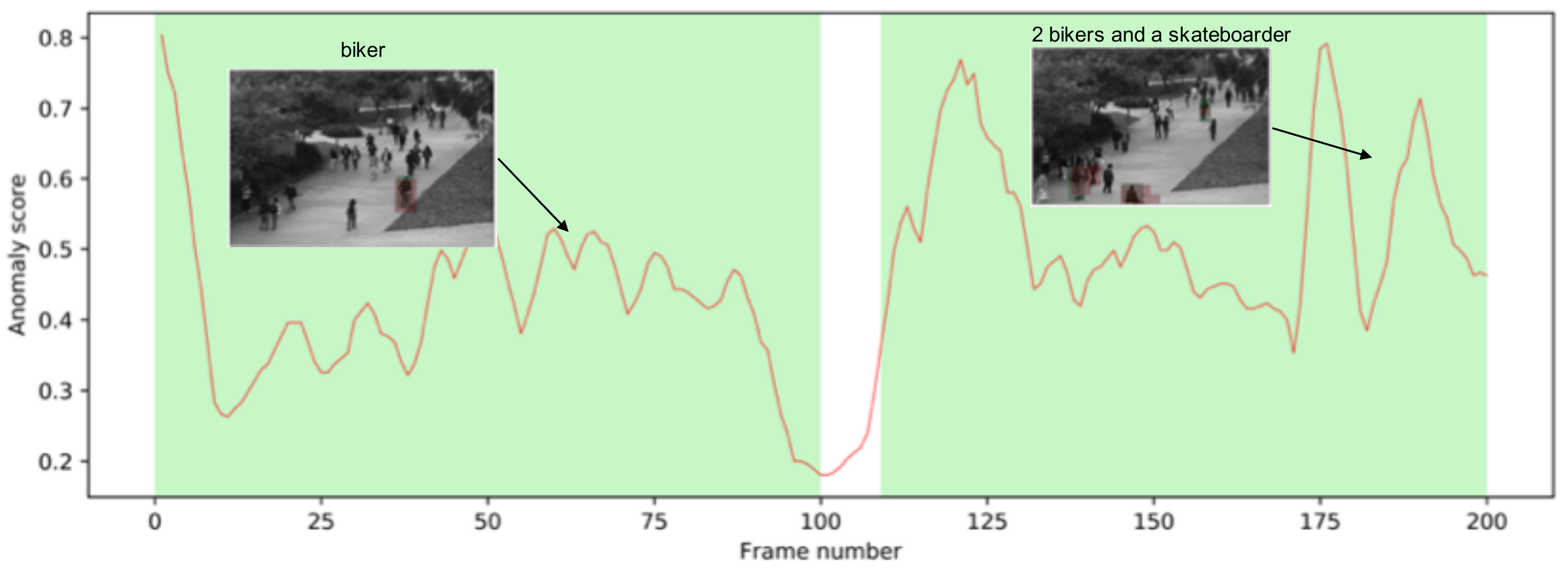}
\caption{Per-frame anomaly score visualization of UCSD Ped1 Test sequence 006.}
\label{fig:ped1_pfe_1}
\end{figure*}

\begin{figure*}
\centering
\includegraphics[width=\linewidth]{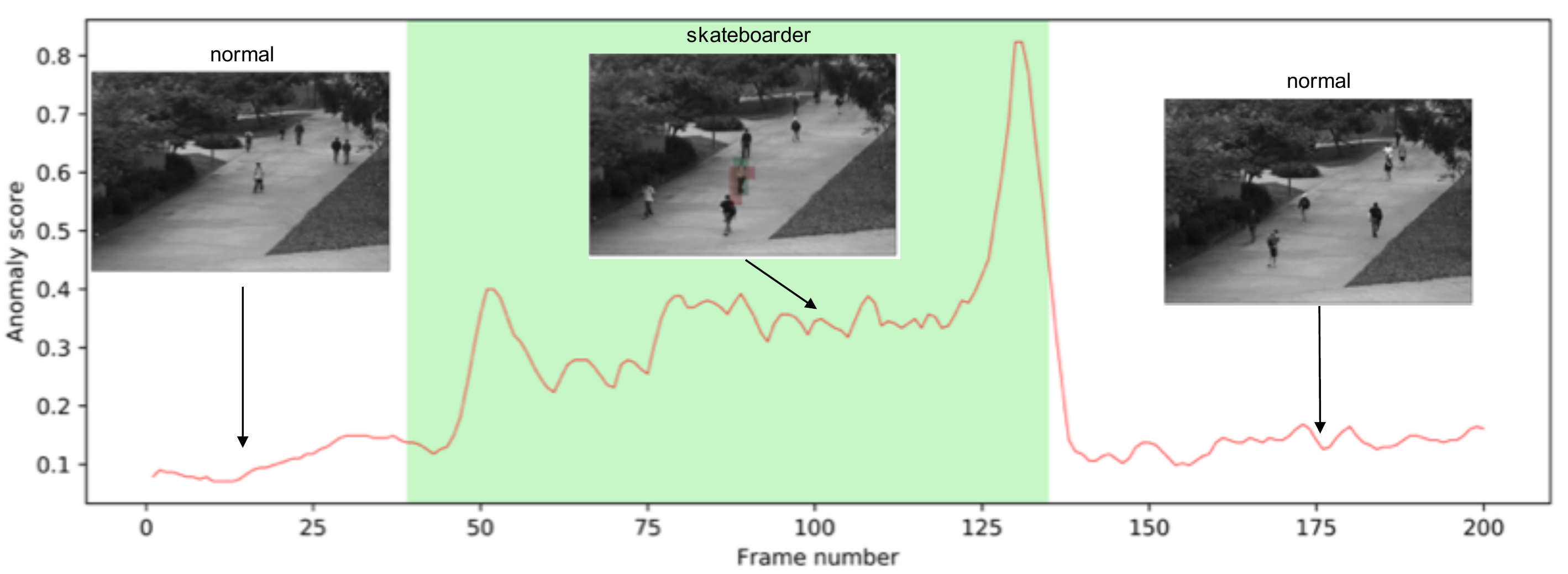}
\caption{Per-frame anomaly score visualization of UCSD Ped1 Test sequence 025.}
\label{fig:ped1_pfe_2}
\end{figure*}

\begin{figure*}
\centering
\includegraphics[width=\linewidth]{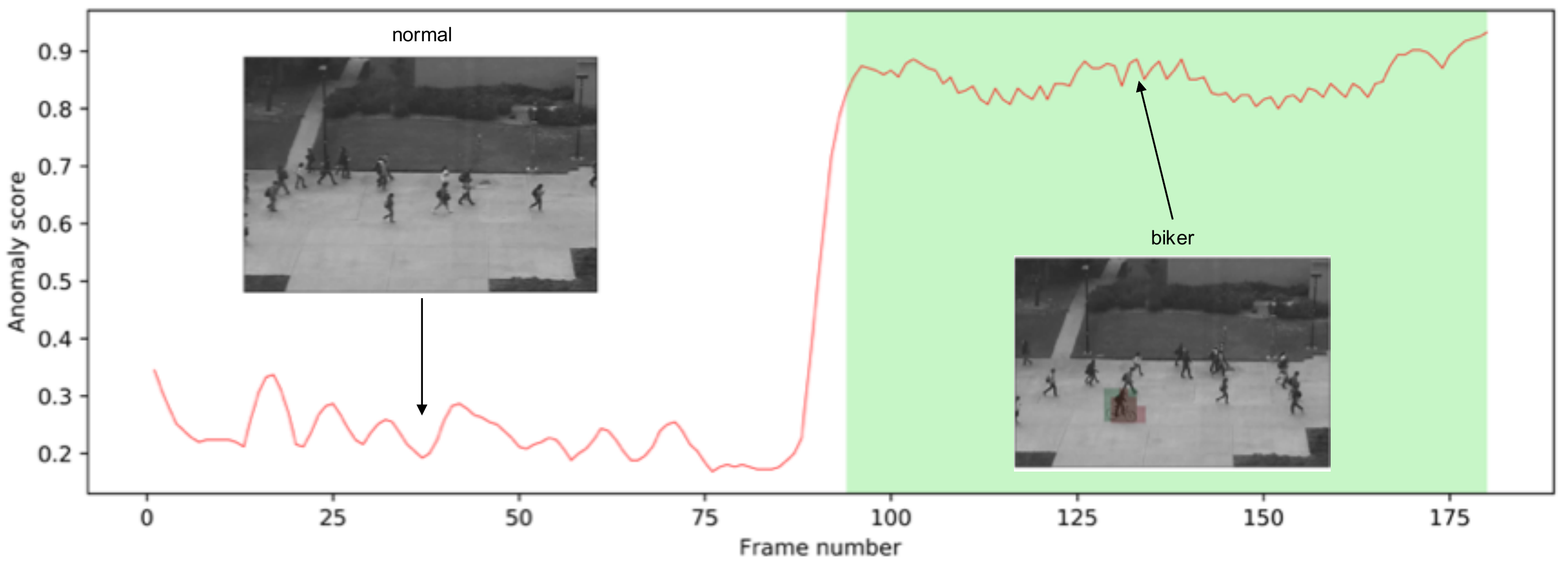}
\caption{Per-frame anomaly score visualization of UCSD Ped2 Test sequence 002.}
\label{fig:ped2_pfe_1}
\end{figure*}

\begin{figure*}
\centering
\includegraphics[width=\linewidth]{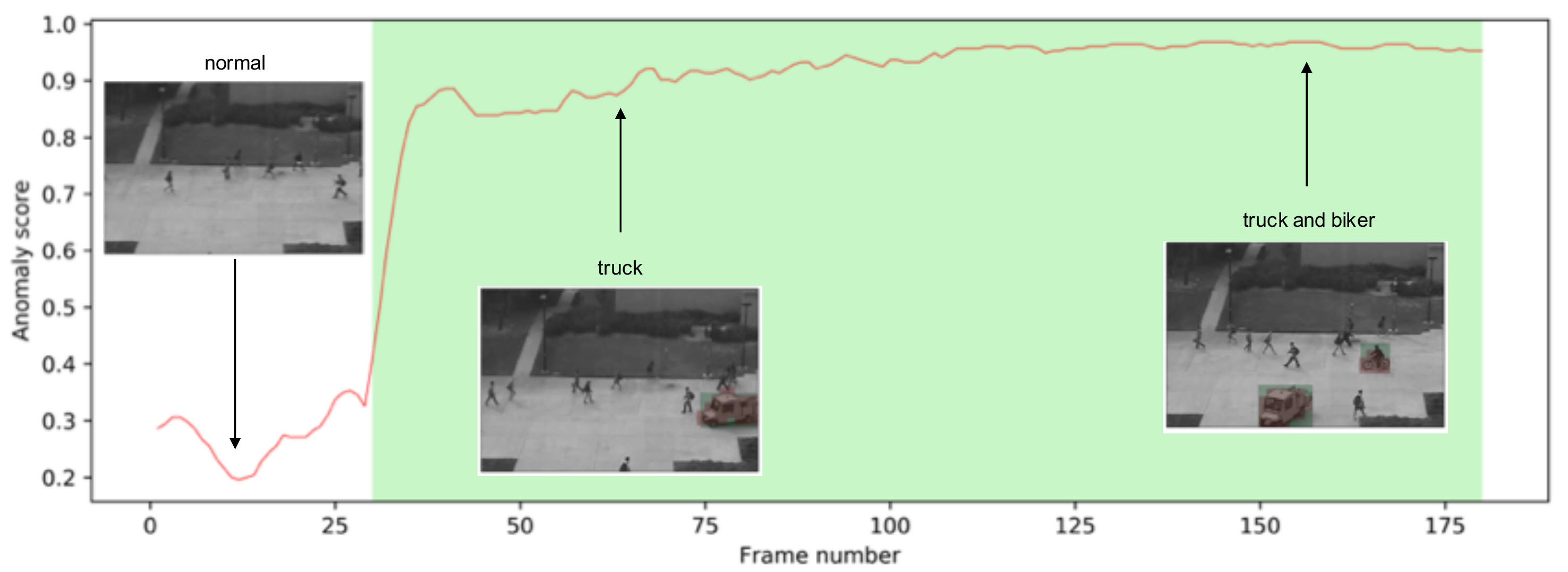}
\caption{Per-frame anomaly score visualization of UCSD Ped2 Test sequence 004.}
\label{fig:ped2_pfe_2}
\end{figure*}

\begin{figure*}
\centering
\includegraphics[width=\linewidth]{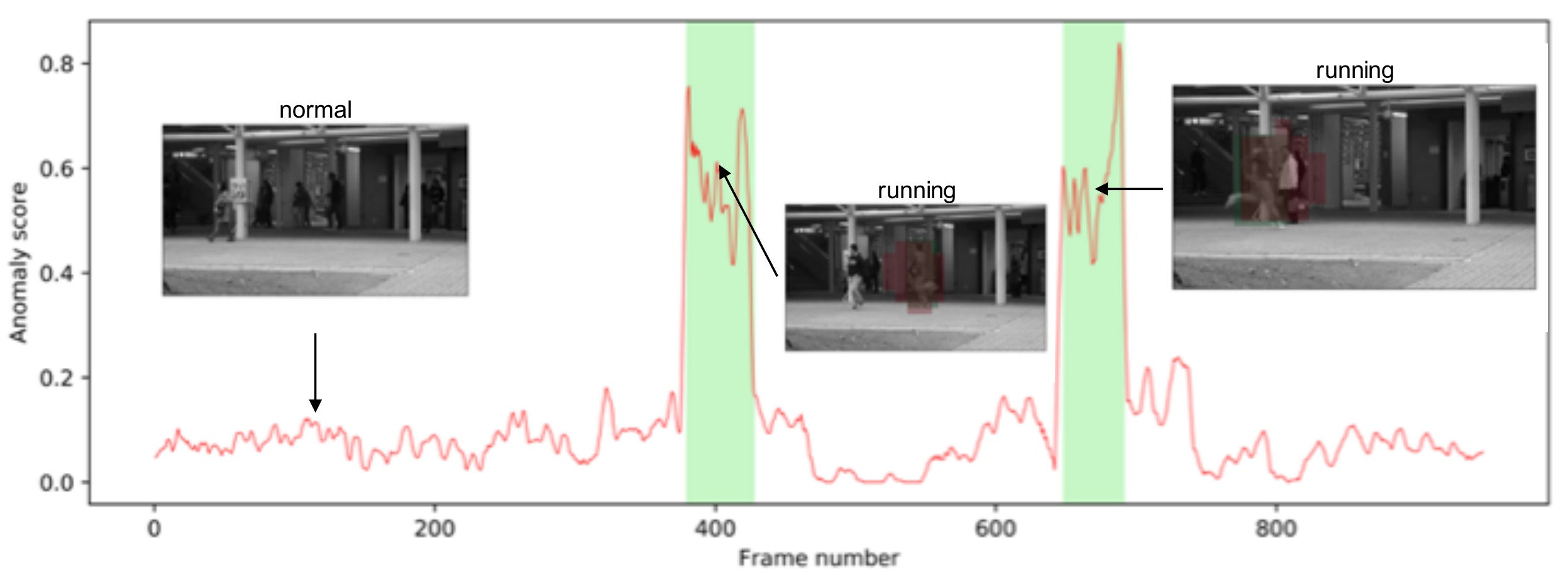}
\caption{Per-frame anomaly score visualization of CUHK Avenue Test sequence 004.}
\label{fig:avenue_pfe_1}
\end{figure*}

\begin{figure*}
\centering
\includegraphics[width=\linewidth]{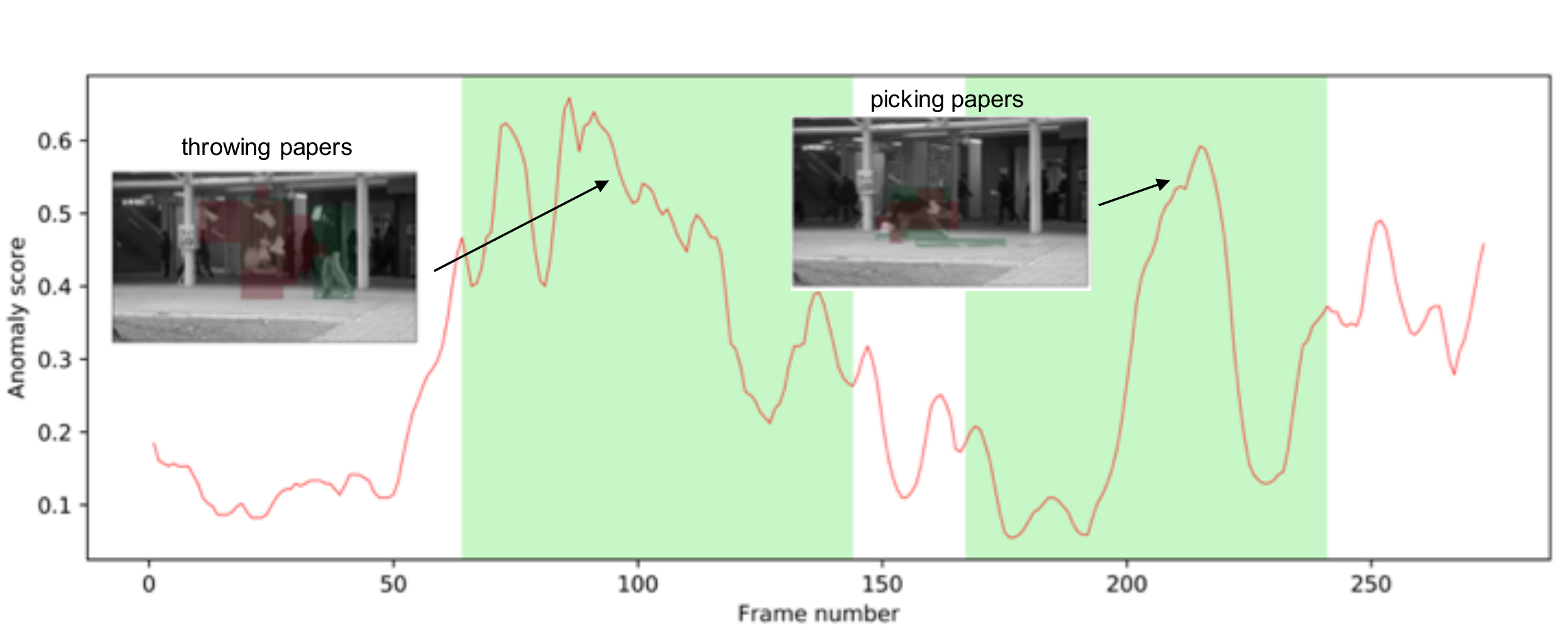}
\caption{Per-frame anomaly score visualization of CUHK Avenue Test sequence 020.}
\label{fig:avenue_pfe_2}
\end{figure*}

Figures \ref{fig:ped1_pfe_1} through \ref{fig:avenue_pfe_2} provide additional frame-level anomaly score visualizations for some test sequences using our approach from all 3 datasets. As in the submission document, green shading on the plot indicates ground truth anomalous frames and we also show detection visualizations at select frames.


\subsection{Visualizations of learned representations for video patch pairs}
\begin{figure*}
\centering
\includegraphics[width=0.5\linewidth]{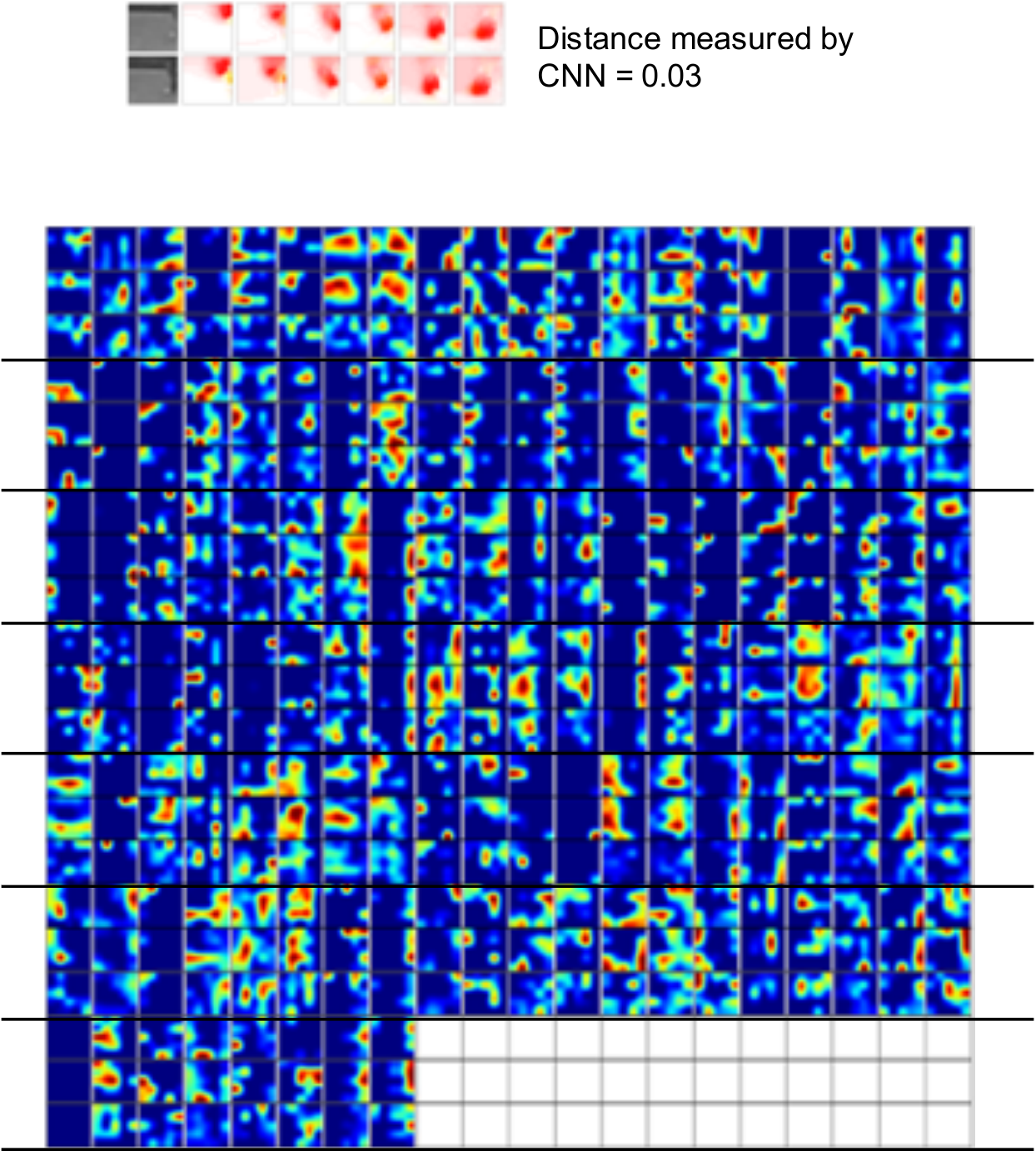}
\caption{Learned representations and their element-wise difference between 2 video patches in UCSD Ped2, visualized.}
\label{fig:activations_1}
\end{figure*}

\begin{figure*}
\centering
\includegraphics[width=0.5\linewidth]{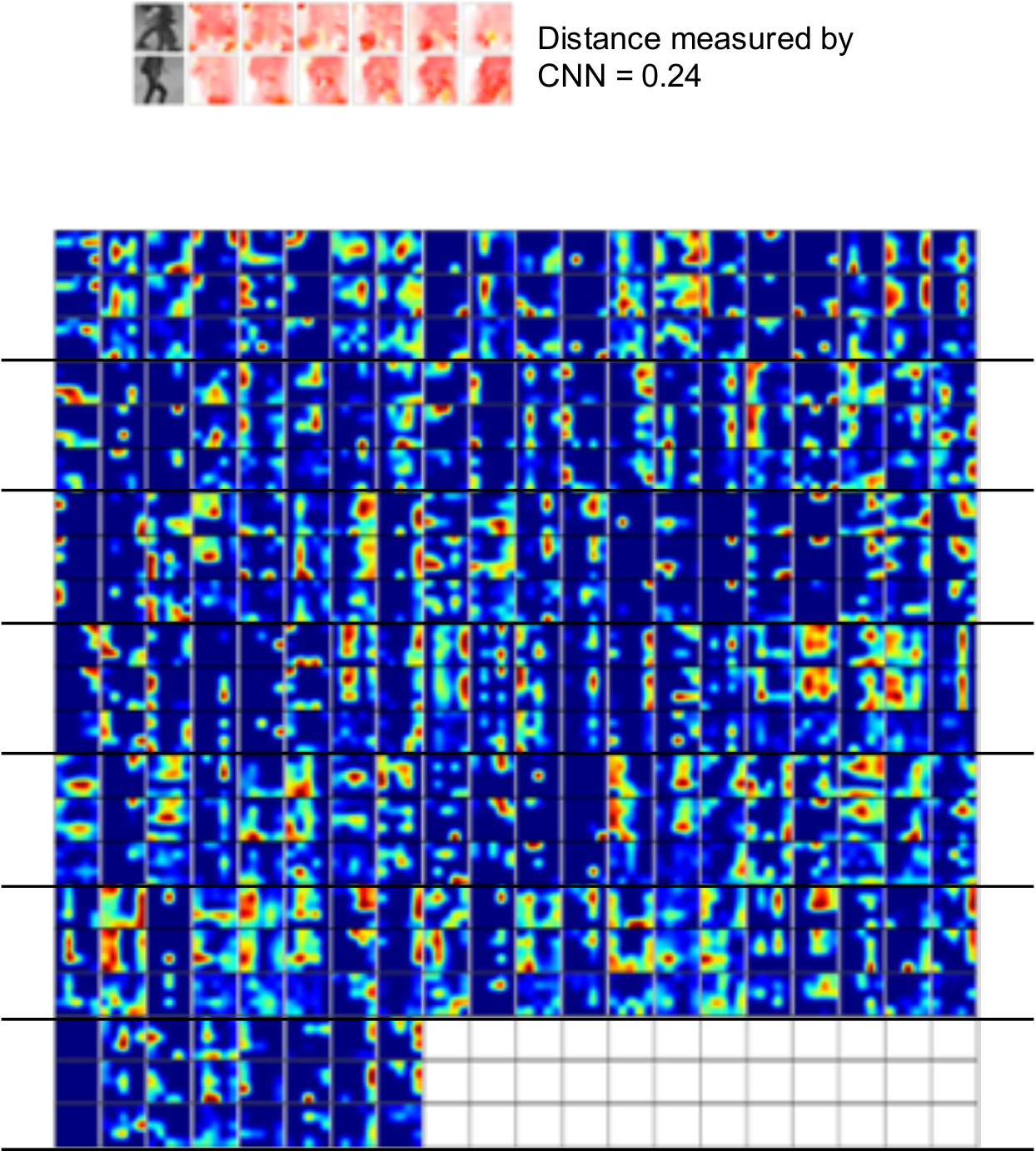}
\caption{Learned representations and their element-wise difference between 2 video patches in UCSD Ped2, visualized.}
\label{fig:activations_4}
\end{figure*}

\begin{figure*}
\centering
\includegraphics[width=0.5\linewidth]{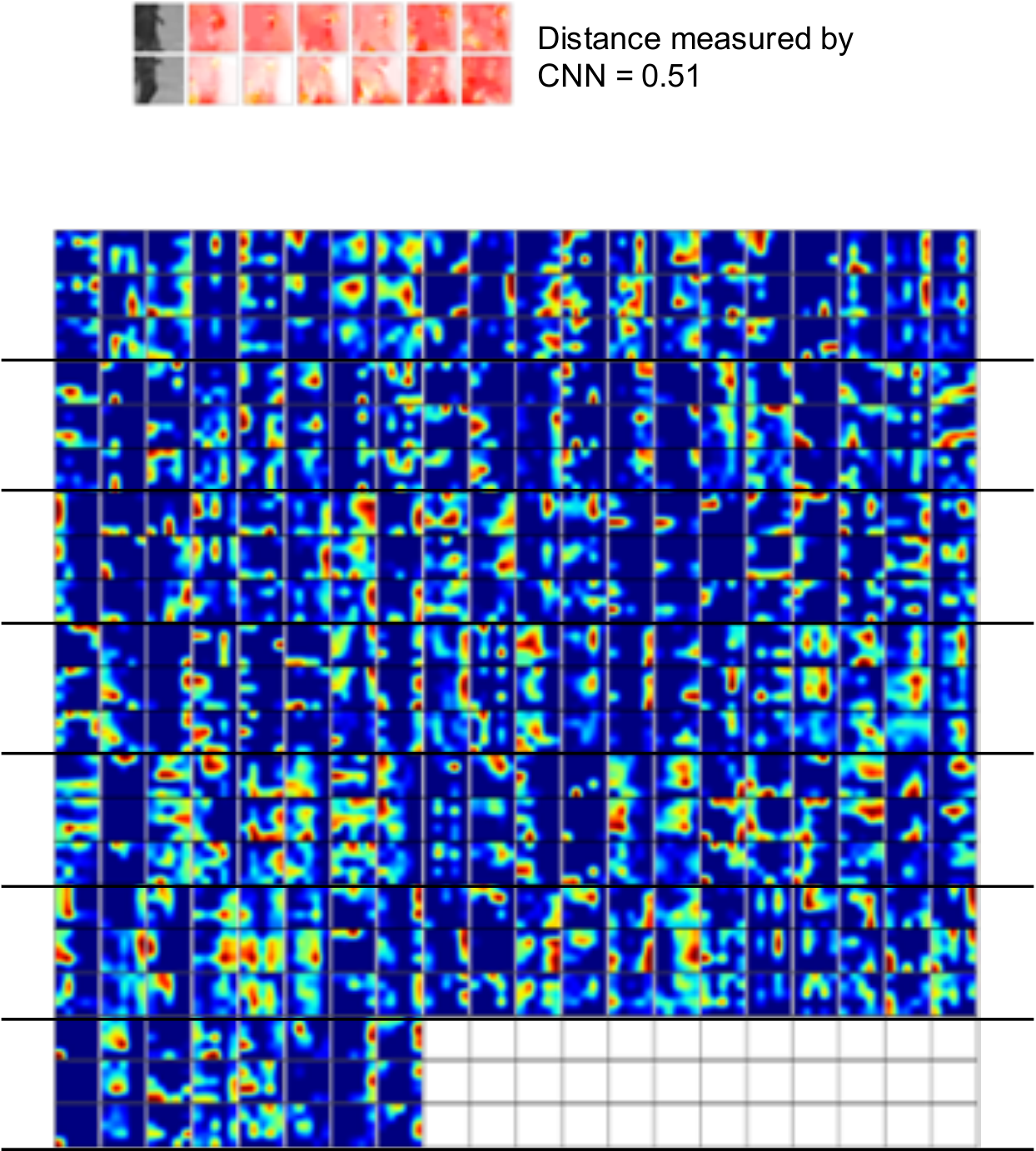}
\caption{Learned representations and their element-wise difference between 2 video patches in UCSD Ped2, visualized.}
\label{fig:activations_6}
\end{figure*}

\begin{figure*}
\centering
\includegraphics[width=0.5\linewidth]{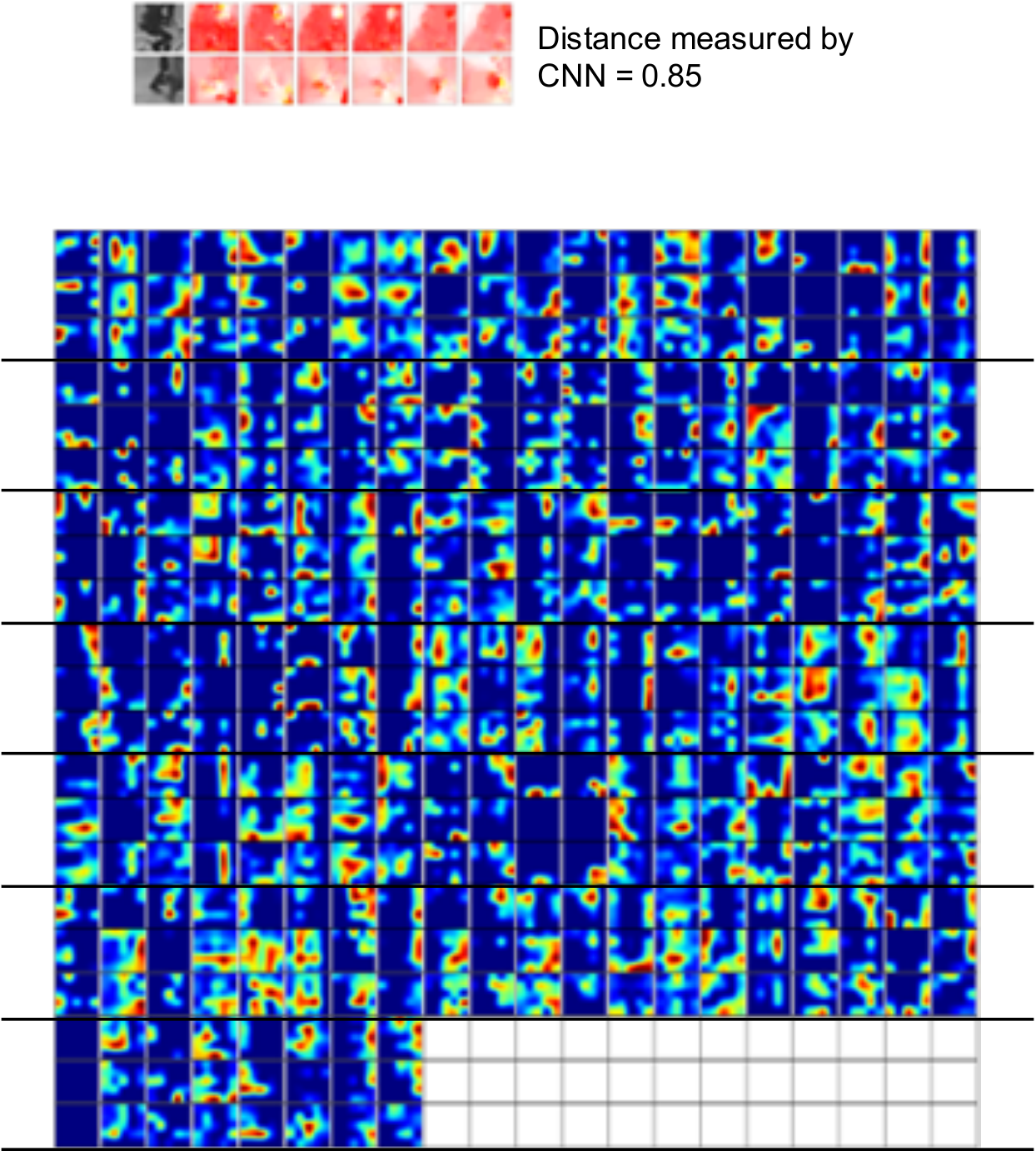}
\caption{Learned representations and their element-wise difference between 2 video patches in UCSD Ped2, visualized.}
\label{fig:activations_7}
\end{figure*}

\begin{figure*}
\centering
\includegraphics[width=0.5\linewidth]{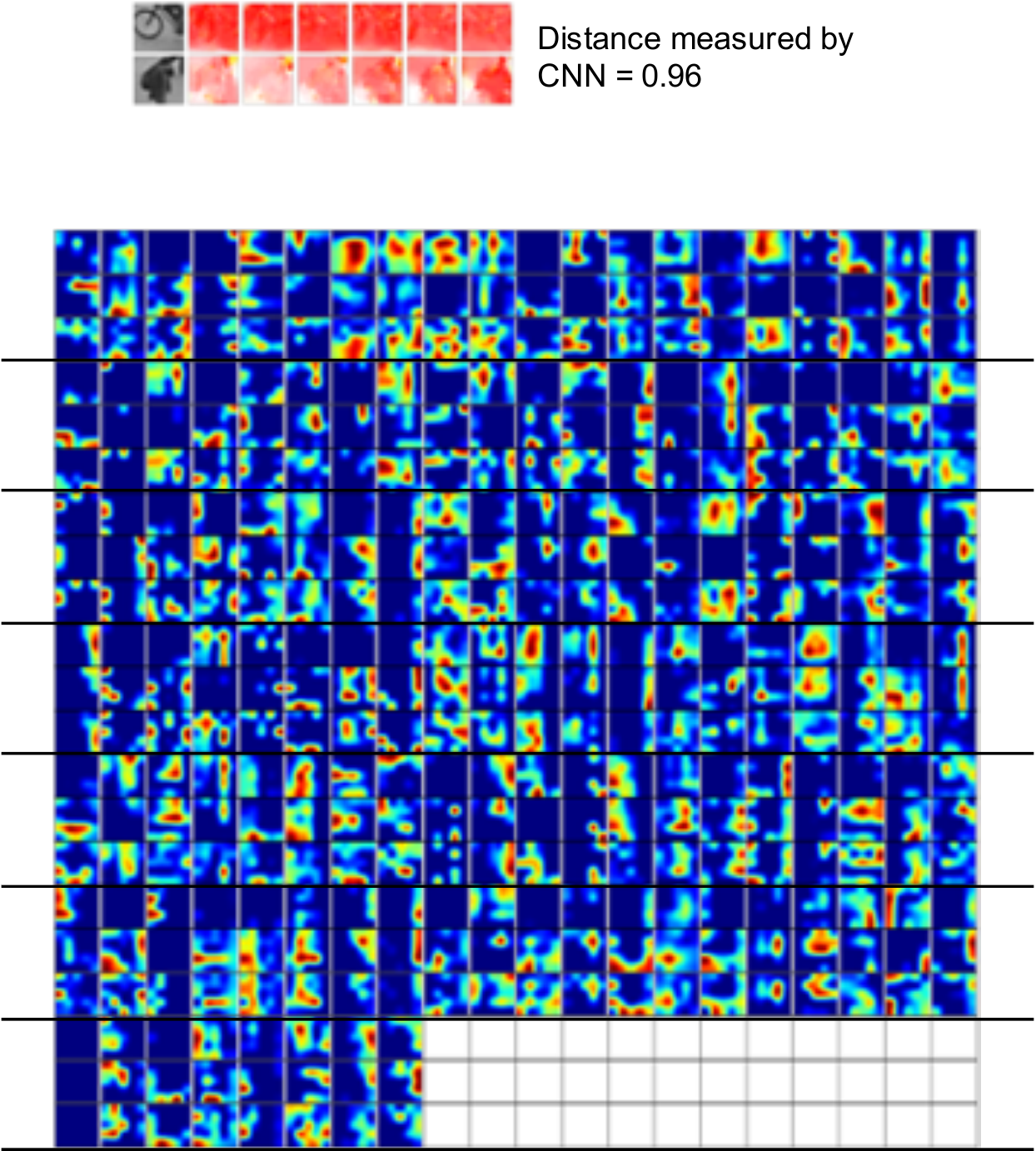}
\caption{Learned representations and their element-wise difference between 2 video patches in UCSD Ped2, visualized.}
\label{fig:activations_8}
\end{figure*}

Figures \ref{fig:activations_1} through \ref{fig:activations_8} show select video patch pairs from UCSD Ped2, their learned representations and the distance measured between them by our CNN. To generate this set of figures, we used the CNN corresponding to the scenario where the target dataset was UCSD Ped2 to give a realistic idea of distance measurement at `test time'. Each group of 3 rows is a visualization of the feature maps of the first video patch before element-wise subtraction (1st row), the second video patch before element-wise subtraction (2nd row), and the element-wise subtraction layer's output (3rd row). All 128 feature maps are shown on columns, wrapping around to the next row when necessary. Specific feature maps could exhibit high activations for features such as speed, direction, velocity, shape, texture and illumination among others.

\end{document}